\documentclass[11pt,letterpaper]{article}
\usepackage[T1]{fontenc}
\usepackage{times}
\usepackage[margin=1in]{geometry}
\usepackage{amsmath,amsfonts,bm}

\def\eqref#1{equation~\ref{#1}}

\def\1{\bm{1}}

\DeclareMathAlphabet{\mathsfit}{\encodingdefault}{\sfdefault}{m}{sl}
\SetMathAlphabet{\mathsfit}{bold}{\encodingdefault}{\sfdefault}{bx}{n}

\usepackage{amssymb}
\usepackage{graphicx,booktabs,multirow,wrapfig,tabularx,array,capt-of}
\usepackage{xcolor}
\usepackage{algorithm,algpseudocode}
\usepackage[authoryear,round]{natbib}
\setcitestyle{citesep={;},aysep={,},yysep={;}}
\usepackage[font=small,labelfont=bf]{caption}
\usepackage{titlesec}
\usepackage{float,placeins}
\titleformat{\section}{\large\bfseries}{\thesection}{0.75em}{}
\titleformat{\subsection}{\normalsize\bfseries}{\thesubsection}{0.75em}{}
\titleformat{\subsubsection}{\normalsize\bfseries}{\thesubsubsection}{0.75em}{}
\titlespacing*{\section}{0pt}{2.0ex plus 0.5ex minus 0.2ex}{1.0ex}
\titlespacing*{\subsection}{0pt}{1.7ex plus 0.4ex minus 0.2ex}{0.8ex}
\titlespacing*{\subsubsection}{0pt}{1.4ex plus 0.3ex minus 0.2ex}{0.6ex}
\usepackage[hidelinks,unicode]{hyperref}
\usepackage{bookmark}
\hypersetup{pdftitle={Frame-Level Temporal Alignment for Human-to-Robot Visual Adaptation},pdfauthor={Xizhe Zhang; Jingfeng Zhang; Zirun Zhou; Hong Jia},pdfsubject={Human-to-robot visual representation adaptation},pdfkeywords={robot learning, temporal alignment, visual representations, domain adaptation}}
\title{\LARGE\bfseries Frame-Level Temporal Alignment for\\Human-to-Robot Visual Adaptation}
\author{Xizhe Zhang\textsuperscript{1}\quad Jingfeng Zhang\textsuperscript{2}\quad Zirun Zhou\textsuperscript{1}\quad Hong Jia\textsuperscript{1}\\[5pt]
\normalsize\textsuperscript{1}The University of Auckland\quad\textsuperscript{2}Fudan University\\[3pt]
\small\href{mailto:xzha593@UoA.auckland.ac.nz}{\nolinkurl{xzha593@UoA.auckland.ac.nz}}\quad\href{mailto:jingfeng.zhang9660@gmail.com}{\nolinkurl{jingfeng.zhang9660@gmail.com}}\\[-1pt]
\small\href{mailto:zzho690@aucklanduni.ac.nz}{\nolinkurl{zzho690@aucklanduni.ac.nz}}\quad\href{mailto:hong.jia@auckland.ac.nz}{\nolinkurl{hong.jia@auckland.ac.nz}}}
\date{}
\begin{document}
\raggedbottom
\maketitle
\vspace{-1.5em}

\begin{abstract}Transferring visual representations pretrained on human videos to robot manipulation requires learning reliable correspondences between human and robot demonstrations. However, paired demonstrations can differ in execution rate and in the proportion of non-key frames that do not directly reflect task progress. Frames at the same relative timestamp may therefore represent different task stages, which can cause correspondence learning to fail. To address these issues, we propose Frame-Level Temporal Alignment (\textbf{FLTA}), a framework that uses two temporal priors to adapt visual encoders pretrained on human videos for robot manipulation. It learns shared task-progress representations without frame-level correspondence annotations, allowing frames at different relative temporal positions to match. A global progress prior combines normalized temporal positions with visual similarity to construct a soft correspondence target. A local temporal order prior penalizes backward transitions while allowing stays and varying forward rates to accommodate execution-rate differences. With ResNet-50 and ViT encoders, our method achieves relative improvements of $46.93\%$ and $65.96\%$, respectively, over the best baselines in average simulation success rates and achieves higher task success rates on real-world manipulation tasks. These results also suggest that effective human--robot adaptation depends less on the number of parameters updated than on which parameters are selected. Our project page is available at \url{https://rtx5090ultra.github.io/FLTA-Project-Page/}.
\end{abstract}

\section{Introduction}\label{sec:intro}

Vision-based robotic manipulation relies on visual representations that capture task-relevant state information for action prediction. Learning such representations solely from robot demonstrations is challenging because robot data are costly to collect and often cover a limited range of tasks. Human videos provide a scalable, complementary source of interaction data, as they are relatively inexpensive to obtain, widely available, and rich in object manipulation experience~\citep{Grauman_2022_CVPR,bahl2023affordances,pmlr-v235-li24cr,Srirama-RSS-24,zhang2026provable}. Pretraining on human videos can encode these experiences into transferable, task-relevant visual representations that support downstream policy learning and generalization beyond task-specific robot demonstrations~\citep{pmlr-v205-nair23a,radosavovic2023real,pmlr-v202-ma23b,pmlr-v235-li24cr}.

However, differences in movement capabilities and appearance between humans and robots create a substantial domain gap between human videos and robot observations~\citep{Zhou_2025_CVPR,ben2006analysis,ganin2016domain}. As such, visual representations need to distinguish task progress because the downstream policy selects its next action based on the current observation~\citep{pmlr-v205-nair23a,Srirama-RSS-24}.
HRAlign~\citep{Zhou_2025_CVPR}, the state-of-the-art solution, uses paired human-robot videos for video-level contrastive alignment, learning shared task semantics to reduce this domain gap. However, frames sampled from paired human and robot videos may show different task stages despite sharing the same task instruction. Aligning their aggregated video representations does not explicitly establish which human and robot frames correspond to similar task progress.

\begin{figure*}[t]
    \centering
    \includegraphics[width=\textwidth]{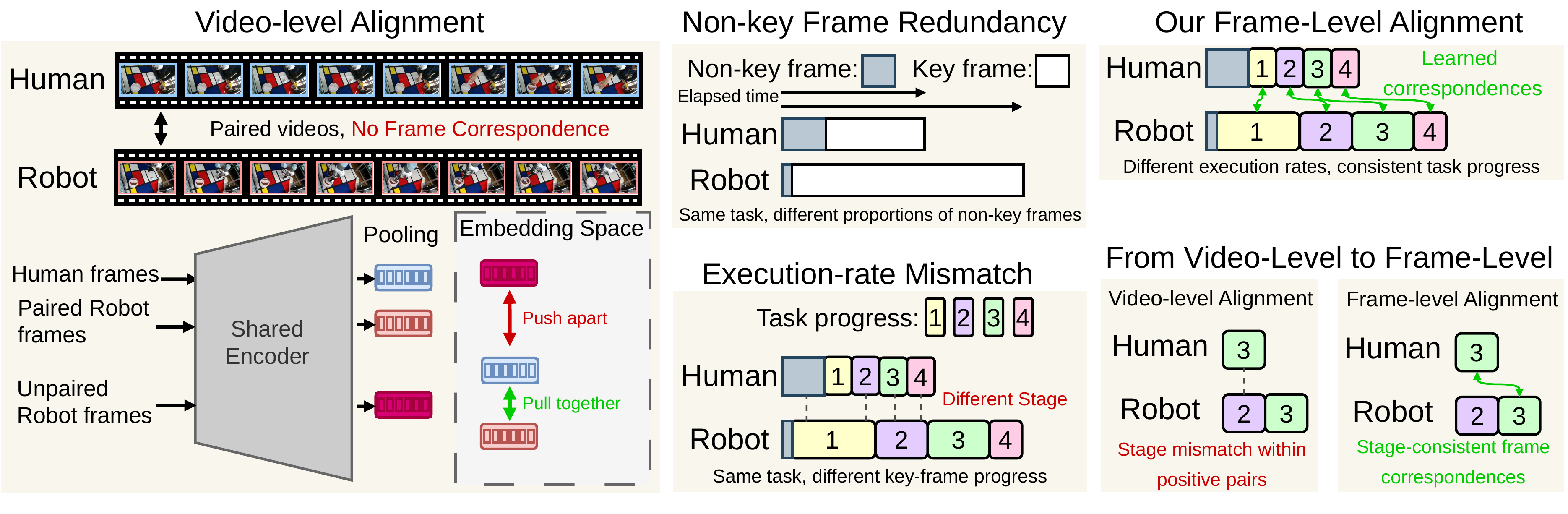}
    \caption{From video-level to frame-level alignment. Video-level alignment brings human and robot video representations closer without explicitly establishing frame-level correspondences. Differences in execution rates and the proportion of non-key frames mean that frames sampled at the same relative time may show different task stages. Our method addresses these temporal discrepancies by learning cross-domain frame correspondences without requiring frame-level annotations.}
    \label{fig:motivation figure}
\end{figure*}

This distinction is apparent for paired datasets such as RH20T~\citep{fang2024rh20t}, which provides human and robot demonstrations of the same task. As shown in Figure~\ref{fig:motivation figure}, these videos exhibit two temporal differences. First, the videos contain different proportions of non-key frames, which do not directly reflect task progress. As a result, the action starts at different relative positions in the two videos. Second, execution-rate mismatch changes the duration of each task stage. Frames sampled at the same relative time may show different task progress. Statistics across tasks and video pairs are provided in Appendix~\ref{app:temporal_differences}. For example, sampled human non-key frames and robot key frames may contribute to a video-level positive pair. The aggregated video representations are pulled together even though the sampled frames show different task progress. Such stage mismatch may reduce the strength of the alignment signal used to learn task-progress representations, making it harder for downstream policies to select appropriate actions.

To address these issues, we propose a frame-level temporal alignment framework that adapts pretrained visual encoders using paired human and robot videos, without frame-level correspondence annotations. As shown in Figure~\ref{fig:motivation figure}, we guide correspondence learning with the inductive bias that \textit{human and robot demonstrations generally progress toward task completion over time}~\citep{Hadji_2021_CVPR,pmlr-v164-zakka22a}. First, we introduce a \textbf{global progress prior $D$} that combines normalized temporal positions with visual similarity to construct a soft correspondence target between human and robot frames. It allows each frame to match multiple candidate frames while penalizing large differences in normalized temporal position, rather than requiring same-position matches. Second, we introduce a \textbf{local temporal order prior $Q$} to guide the order of matches for consecutive sampled frames.
As time progresses from the current human frame to the next sampled human frame, $Q$ encourages the corresponding robot match to stay or move forward by penalizing backward transitions.
This encourages temporal order while allowing for execution-rate differences. During adaptation, we update only the affine parameters of normalization layers, a restricted parameterization previously used for domain adaptation~\citep{wang2020tent}. This restricts the parameter search space during adaptation, with the aim of improving the visual representations for downstream robotic manipulation.

On AGNOSTOS~\citep{zhou2026exploring}, our method achieves a $48.27\%$ relative improvement in average success rate over HRAlign with R3M-RN50~\citep{pmlr-v205-nair23a} initialization. With D4R-ViT~\citep{pmlr-v229-dasari23a} initialization, it achieves a $65.96\%$ relative improvement over HRP~\citep{Srirama-RSS-24}. With D4R-ViT initialization, it also improves success rates on all four real-world manipulation tasks compared with the Frozen and HRP baselines. Our experiments further show that effective human--robot adaptation does not require a large number of trainable parameters. Updating only 53.1K normalization-affine parameters achieves higher success rates than the late-adapter variant, despite using approximately $29.8\times$ fewer trainable parameters. Adaptation only takes approximately 13 hours on a single NVIDIA RTX 5090 GPU, without multi-GPU training. These results suggest that effective human--robot adaptation depends less on the number of parameters updated than on which parameters are selected.

\section{Human-to-Robot Domain Adaptation}\label{sec:Preliminaries and Methodology}
We formulate the human-to-robot domain gap as a representation adaptation problem. For paired human and robot videos of the same task with different temporal structures, video-level alignment does not explicitly establish frame-level correspondences, as illustrated in Figure~\ref{fig:motivation figure}. We therefore adapt a pretrained visual encoder through frame-level temporal alignment without frame-level correspondence annotations. The adapted encoder is then frozen for downstream robot policy learning.

\subsection{Problem Formulation}\label{subsec:problem_formulation}
Let $\mathcal D=\{(L_\kappa,\mathcal E_\kappa)\}_{\kappa=1}^{K}$ denote a paired human--robot demonstration dataset, where $K$ is the number of tasks. Here, $L_\kappa$ is the instruction for task $\kappa$, and $\mathcal E_\kappa$ contains its human--robot video pairs. For each pair $(\mathcal H,\mathcal R)\in \mathcal E_\kappa$, $\mathcal H$ and $\mathcal R$ denote the human and robot demonstrations, respectively. The videos may differ in temporal structure, and no frame-level correspondence annotations are provided. For downstream policy learning, we train a policy to predict robot actions from robot visual observations. In the human-video pretraining paradigm, a visual encoder $f_\theta$ is first pretrained on human videos. The policy then takes the encoded representation $f_\theta(O)$, rather than the raw robot observation $O$, as its visual state input. To handle the human-robot domain gap, we adapt $f_\theta$ using $\mathcal D$. The adapted encoder provides visual representations better suited to robot observations $O$, helping the downstream policy predict the next action more reliably.

\subsection{Framework Overview}\label{subsec:Framework Overview}

\begin{figure*}[t]
    \centering
    \includegraphics[width=\textwidth]{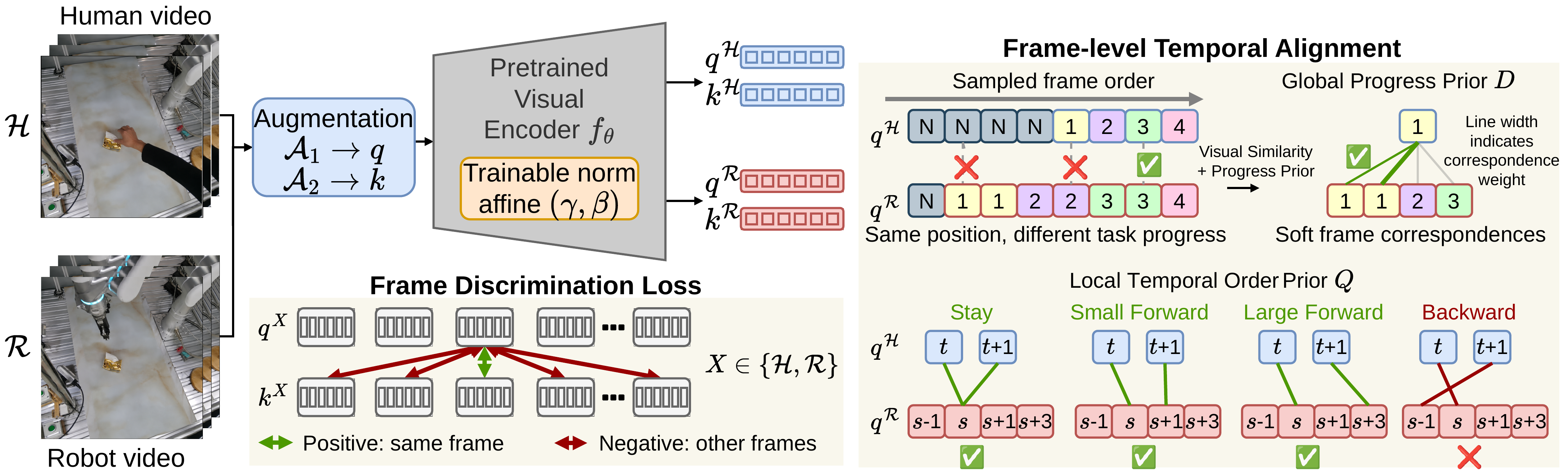}
    \caption{Overview of the proposed temporal alignment framework. Here, only the normalization-affine parameters of the visual encoder are updated. The auxiliary frame discrimination loss learns discriminative frame representations within each domain. The global progress prior $D$ and visual similarity jointly construct soft correspondences. The local temporal order prior $Q$ penalizes backward transitions while allowing stays and forward transitions of varying magnitudes. Numbered stages are shown only for illustration.}
    \label{fig:ours_framework}
\end{figure*}

As shown in Figure~\ref{fig:ours_framework}, our framework learns frame-level temporal correspondences using visual similarity, a global progress prior, and a local temporal order prior. An auxiliary frame discrimination loss supports cross-domain matching.

\subsection{Frame-Level Temporal Alignment}\label{subsec:soft alignment}
Motivated by prior work on self-supervised temporal correspondence learning~\citep{Dwibedi_2019_CVPR,Haresh_2021_CVPR,Liu_2022_CVPR}, we learn correspondences between paired human and robot videos without frame-level annotations. At each iteration, we uniformly sample $|\mathcal T|$ frames from each full video clip, including non-key frames, and retain their temporal order. Here, $\mathcal T$ denotes the sampled temporal index set, with the same number of frames $|\mathcal T|=8$ sampled from each video by default. We index the sampled human and robot frames by $t,s\in\{0,\ldots,|\mathcal T|-1\}$, respectively, using their positions within the sampled sequences rather than their original video frame indices. Using these sampled sequences, we compute cross-domain visual similarities and guide correspondence learning with two temporal priors, supported by an auxiliary frame discrimination loss.

\noindent\textbf{Visual Similarity. }
The encoder $f_\theta$ maps sampled human and robot frames to representations $q_t^{\mathcal H}$ and $q_s^{\mathcal R}$, respectively. We then construct a cross-domain visual similarity matrix $S$, where $S_{ts}=\operatorname{sim}(q_t^{\mathcal H},q_s^{\mathcal R})$ is the cosine similarity between the two frame representations. We convert the similarity matrix $S$ into the matching probability matrix $P$ by applying a row-wise softmax, where $P_{ts}$ is the predicted probability of matching human frame $t$ to robot frame $s$.

\noindent\textbf{Global Progress Prior $D$. }
As illustrated in Figure~\ref{fig:ours_framework}, human and robot frames at the same sampled position can represent different task progress. We therefore use soft correspondences across sampled positions, guided by both visual similarity and temporal position. We define $D_{ts}=|t-s|/(|\mathcal T|-1)$, so larger differences between the sampled frame indices $t$ and $s$ incur higher costs. Combining this prior with the visual matching cost $1-S$ gives the soft correspondence target:
\begin{equation}
    T = \operatorname{Sinkhorn} \left(\exp\left[ -\frac{ 1-S+\rho D }{ \epsilon }\right]\right).
    \label{eq:global_progression_teacher}
\end{equation}

Sinkhorn normalization converts the combined costs into correspondence weights~\citep{cuturi2013sinkhorn,pmlr-v97-su19b}. Each row of $T$ assigns weights to candidate robot frames for one human frame. Here, $\rho$ controls the weight of $D$, while larger $\epsilon$ values spread the weights more evenly across candidates.

\noindent\textbf{Local Temporal Order Prior $Q$. }
We introduce a local temporal order prior $Q$ to penalize backward transitions while allowing different execution rates. Here, $P_{ts}$ gives the predicted probability of matching human frame $t$ to robot frame $s$. For consecutive sampled human frames $t$ and $t+1$, let $s$ and $s'$ denote their candidate correspondence positions in the robot sequence, respectively. Within the sampled robot sequence, $s'$ may equal $s$, move forward to $s+1,s+2,\ldots$, or move backward to $s-1,s-2,\ldots$. We assign the following cost to each transition from $s$ to $s'$:
\begin{equation}
    Q_{ss'} = \left( \frac{\max(s-s',0)}{|\mathcal T|-1} \right)^2.
    \label{eq:local_transition_prior}
\end{equation}
This cost penalizes backward transitions ($s'<s$), with larger backward distances receiving higher penalties. Stays and forward transitions are not penalized by $Q$.

\noindent\textbf{Bidirectional Alignment Objective. }
We train $P$ to match the soft target $T$ using cross-entropy and penalize backward transitions using $Q$. For human-to-robot alignment, the loss is:
\begin{equation}
    \mathcal L_{\mathrm{TA}}^{\mathcal H\rightarrow\mathcal R} = \operatorname{CE}(T,P)+\mathcal R_Q(P).
    \label{eq:bidirectional_alignment}
\end{equation}

Here, $\operatorname{CE}$ denotes the mean row-wise cross-entropy. We recompute $T$ at each iteration but do not propagate gradients through it. The regularizer $\mathcal R_Q(P)$ discourages backward transitions in the predicted correspondences. We exchange the human and robot sequences to compute the reverse loss, then average the two directional losses to obtain $\mathcal L_{\mathrm{TA}}$. Detailed definitions are provided in Appendix~\ref{app:alignment_objective}.

\noindent\textbf{Frame Discrimination Loss.}
We use an auxiliary InfoNCE loss~\citep{oord2018representation}, denoted by $\mathcal L_{\mathrm{InfoNCE}}$, to learn discriminative frame representations for cross-domain alignment. Within each domain, augmented views of the same frame form positive pairs, while other frames in the same domain and batch serve as negatives. The full formulation is provided in Appendix~\ref{app:frame_discrimination}.

\noindent\textbf{Overall Training Objective. }
The overall training objective is $\mathcal L_{\mathrm{total}} = \lambda_1\mathcal L_{\mathrm{InfoNCE}} + \lambda_2 \mathcal L_{\mathrm{TA}}$, where $\lambda_1$ and $\lambda_2$ control the contributions of the two loss functions, respectively. We minimize this loss by updating only the affine parameters of the encoder's normalization layers; all other encoder parameters remain fixed.

\section{Experiments}\label{sec:exp}
We aim to answer four questions. \textbf{(Q1)} Does frame-level temporal alignment improve the success rates of downstream manipulation policies in simulation and the real world? \textbf{(Q2)} Do the adapted visual representations achieve higher human-to-robot task-progress retrieval accuracy than HRP and HRAlign? \textbf{(Q3)} How do frame discrimination, the global progression prior, and local transition regularization contribute to performance, and how does our objective compare with alternatives? \textbf{(Q4)} How does performance vary across encoder initializations, adaptation strategies, and paired-data budgets? The results show improved manipulation success and task-progress retrieval accuracy, with ablations supporting the contributions of $D$ and $Q$. Our method outperforms HRAlign under the same late-adapter configuration. For R3M-RN50, BN-affine fine-tuning achieves a higher success rate than our late-adapter variant with only 53.1K trainable encoder parameters.

\begin{figure*}[t]
    \centering
    \includegraphics[width=\textwidth]{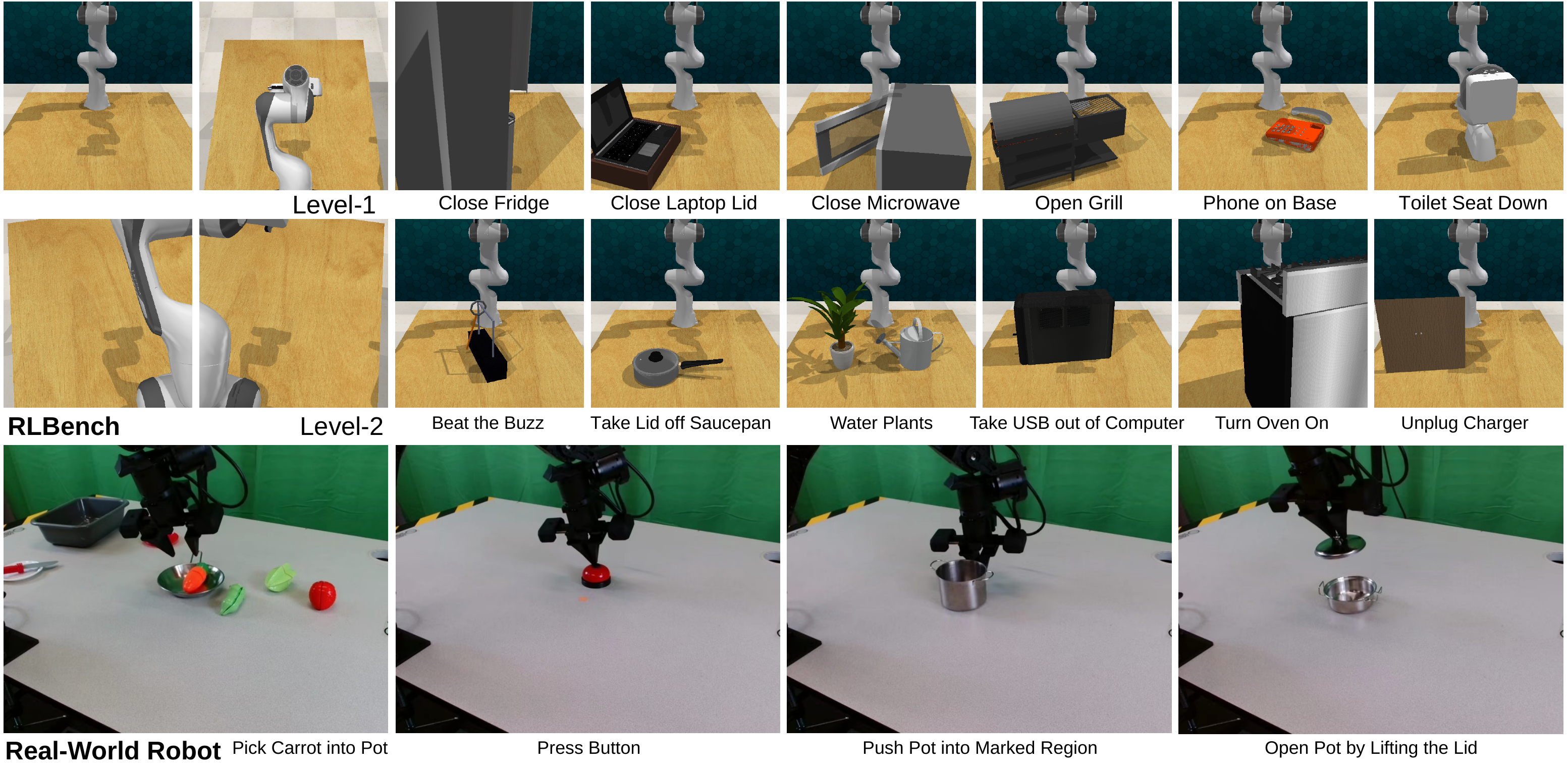}
    \caption{Simulated and real-world evaluation environments. Top: Level 1 and Level 2 unseen manipulation tasks from the AGNOSTOS benchmark in RLBench. Bottom: Four real-world tasks used on the Trossen WidowX AI Arm.}
    \label{fig:eval environments}
\end{figure*}

\noindent\textbf{Experimental Setups. }
Figure~\ref{fig:eval environments} shows the simulated and real-world evaluation environments.

\textbf{In simulation}, we evaluate cross-task zero-shot generalization using the AGNOSTOS benchmark~\citep{zhou2026exploring} in RLBench~\citep{james2020rlbench}. Policies are trained on 18 task categories and evaluated on 23 unseen categories, with no overlap between the training and testing categories. Level-1 contains 13 tasks that share some object or action semantics with the training tasks. Level-2 contains 10 tasks with new object and action semantics, testing transfer beyond these familiar semantics. The complete task split is listed in Table~\ref{tab:agnostos_tasks} in the appendix. Following HRAlign's use of RVT for downstream manipulation~\citep{Zhou_2025_CVPR,goyal2023rvt}, we adopt RVT-2~\citep{goyal2024rvt} as our policy architecture. To evaluate the transferred representations, we insert the frozen visual encoder under evaluation into Stage 2 and reduce the number of policy attention layers in that stage from eight to one. Stage 1 remains unchanged.

\textbf{For real-world evaluation}, we use the Trossen WidowX AI arm for four tasks: \textit{Pick Carrot into Pot}, \textit{Press Button}, \textit{Push Pot into Marked Region}, and \textit{Open Pot by Lifting the Lid}. These tasks cover pick-and-place, button pressing, goal-directed pushing, and grasp-and-lift manipulation. We adopt ACT~\citep{zhao2023learning} through the Trossen LeRobot training pipeline, replacing its visual encoder with the frozen encoder under evaluation. Each task uses 40 training demonstrations and 20 evaluation episodes. Environment and policy configurations are detailed in Appendix~\ref{appendix sec: Downstream Policy Architectures and Hyperparameters}.

\noindent\textbf{Baselines. }
Our model is initialized based on two different encoder types: ViT~\citep{dosovitskiy2020image} and RN50 (ResNet-50)~\citep{He_2016_CVPR}. In embodied learning, these two standard encoders represent different pretraining paths~\citep{vc2023,jing2023exploring}. R3M~\citep{pmlr-v205-nair23a} uses an RN50 encoder to learn task progress over time from first-person human videos through temporal contrastive learning and visual-language alignment. In contrast, the ImageNet branch of D4R~\citep{pmlr-v229-dasari23a} uses a ViT encoder to reconstruct masked patches in static images, but does not include explicit temporal supervision. Thus, the frozen R3M and D4R-ImageNet models were selected as our vanilla baselines.
Our primary adaptation baselines are HRP-ImageNet~\citep{Srirama-RSS-24} and HRAlign~\citep{Zhou_2025_CVPR}. HRP-ImageNet uses 150K human video clips from Ego4D~\citep{Grauman_2022_CVPR}, without robot demonstrations for cross-domain alignment. Our method uses approximately 64K paired human--robot demonstrations from RH20T~\citep{fang2024rh20t}. HRAlign uses the same source dataset as our method but differs in data processing and fine-tuning strategy. To examine adaptation across different pretraining data and objectives, we additionally evaluate DecisionNCE~\citep{pmlr-v235-li24cr} and AcTOL~\citep{zhang2026provable}, both initialized from CLIP and further pretrained on EPIC-KITCHENS-100~\citep{damen2022rescaling}, a human-video corpus with a shorter total duration than Ego4D used by R3M.

\noindent\textbf{Implementation Details.}
We adapt the pretrained encoders using task-level human-robot video pairs from RH20T~\citep{fang2024rh20t}, retaining approximately 64K human--robot video pairs after data filtering. At each training iteration, we uniformly sample eight frames from each video. We set the loss weights to $\lambda_1=0.5$ and $\lambda_2=1$. We set $(\rho,\epsilon)=(0.25,0.025)$ for R3M-RN50, DecisionNCE-CLIP, and AcTOL-CLIP, and $(0.5,0.05)$ for D4R-ViT. Detailed training settings are provided in Appendix~\ref{appendix:pre-training details}.

\subsection{Downstream Language-Conditioned Policy Evaluation}
\begin{table}[t]
    \centering
    \small
    \caption{Zero-shot manipulation success rates (\%) on the 23 unseen AGNOSTOS tasks, with standard deviations in parentheses. Avg.\ denotes the average over all 23 tasks. Bold indicates the best result within each initialization group. Green indicates relative gains over HRP for D4R-ViT and HRAlign for the other initializations. DNCE denotes DecisionNCE. $^{*}$ denotes reproduced.}
    \label{tab:main result}
    \setlength{\tabcolsep}{3pt}
    \renewcommand{\arraystretch}{1.08}

    \resizebox{\linewidth}{!}{%
        \begin{tabular}{@{}llccc@{\hspace{1.5em}}llccc@{}}
            \toprule
            Encoder & Method
            & L1 $\uparrow$ & L2 $\uparrow$ & Avg. $\uparrow$
            & Encoder & Method
            & L1 $\uparrow$ & L2 $\uparrow$ & Avg. $\uparrow$ \\
            \midrule

            \multirow{4}{*}{D4R-ViT}
            & Frozen
            & 8.31 {\tiny (0.50)}
            & 11.60 {\tiny (0.33)}
            & 9.74 {\tiny (0.25)}
            & \multirow{4}{*}{DNCE-CLIP}
            & Frozen
            & 10.05 {\tiny (1.24)}
            & 9.73 {\tiny (1.32)}
            & 9.91 {\tiny (1.14)} \\

            & + HRP
            & 10.56 {\tiny (1.67)}
            & 12.13 {\tiny (0.38)}
            & 11.25 {\tiny (1.03)}
            & & + HRAlign$^{*}$
            & 12.62 {\tiny (2.06)}
            & 9.47 {\tiny (1.05)}
            & 11.25 {\tiny (1.58)} \\

            & \textbf{+ Ours}
            & \textbf{18.05 {\tiny (0.52)}}
            & \textbf{19.47 {\tiny (0.50)}}
            & \textbf{18.67 {\tiny (0.49)}}
            & & \textbf{+ Ours}
            & \textbf{15.38 {\tiny (0.44)}}
            & \textbf{16.13 {\tiny (1.00)}}
            & \textbf{15.71 {\tiny (0.22)}} \\

            & &
            \textcolor{green!50!black}{(+70.93\%)}
            & \textcolor{green!50!black}{(+60.51\%)}
            & \textcolor{green!50!black}{(+65.96\%)}
            & & &
            \textcolor{green!50!black}{(+21.87\%)}
            & \textcolor{green!50!black}{(+70.33\%)}
            & \textcolor{green!50!black}{(+39.64\%)} \\
            \midrule

            \multirow{4}{*}{R3M-RN50}
            & Frozen
            & 12.62 {\tiny (0.25)}
            & 11.73 {\tiny (0.50)}
            & 12.23 {\tiny (0.36)}
            & \multirow{4}{*}{AcTOL-CLIP}
            & Frozen
            & 11.28 {\tiny (1.53)}
            & 11.47 {\tiny (0.19)}
            & 11.36 {\tiny (0.82)} \\

            & + HRAlign
            & 11.59 {\tiny (1.62)}
            & 12.80 {\tiny (0.33)}
            & 12.12 {\tiny (0.78)}
            & & + HRAlign$^{*}$
            & 11.90 {\tiny (1.47)}
            & 12.27 {\tiny (0.19)}
            & 12.06 {\tiny (0.78)} \\

            & \textbf{+ Ours}
            & \textbf{16.82 {\tiny (1.24)}}
            & \textbf{19.47 {\tiny (0.19)}}
            & \textbf{17.97 {\tiny (0.67)}}
            & & \textbf{+ Ours}
            & \textbf{13.23 {\tiny (0.75)}}
            & \textbf{21.73 {\tiny (0.50)}}
            & \textbf{16.93 {\tiny (0.64)}} \\

            & &
            \textcolor{green!50!black}{(+45.13\%)}
            & \textcolor{green!50!black}{(+52.11\%)}
            & \textcolor{green!50!black}{(+48.27\%)}
            & & &
            \textcolor{green!50!black}{(+11.18\%)}
            & \textcolor{green!50!black}{(+77.10\%)}
            & \textcolor{green!50!black}{(+40.38\%)} \\
            \bottomrule
        \end{tabular}%
    }
\end{table}

\noindent\textbf{Simulation results. }
As shown in Table~\ref{tab:main result}, our method achieves the highest average success rates on both L1 and L2 across all four encoder initializations. The overall relative gains are $65.96\%$ over HRP with D4R-ViT and $48.27\%$ over HRAlign with R3M-RN50. For DecisionNCE-CLIP and AcTOL-CLIP, the overall relative gains over HRAlign are $39.64\%$ and $40.38\%$, respectively. Relative gains are larger on L2 for R3M-RN50, DecisionNCE-CLIP, and AcTOL-CLIP, whereas D4R-ViT shows a larger relative gain on L1.
Per-task results in Table~\ref{tab:agnostos_per_task} show that lid opening makes the largest contribution to the improved L2 averages across all four encoder initializations. These gains across D4R, R3M, and CLIP-based initializations suggest that frame-level temporal alignment helps bridge the human--robot domain gap, improving zero-shot transfer of pretrained object and manipulation knowledge.

\begingroup
\setlength{\emergencystretch}{2em}
\begin{wraptable}{r}{0.45\textwidth}
    \vspace{-8pt}
    \centering
    \small
    \caption{Real-world success rates using D4R-ViT initialization. T1--T4 denote pick--place, press, push, and open, respectively.}
    \label{tab:real_robot_results}
    \renewcommand{\arraystretch}{1.08}

    \begin{tabular*}{\linewidth}{@{\extracolsep{\fill}}lccccc@{}}
        \toprule
        Method & T1 & T2 & T3 & T4 & Avg. $\uparrow$ \\
        \midrule
        Frozen & 0.35 & 0.25 & 0.55 & 0.00 & 0.29 \\
        + HRP & 0.40 & 0.50 & 0.35 & 0.05 & 0.33 \\
        \textbf{+ Ours}
        & \textbf{0.65}
        & \textbf{0.60}
        & \textbf{0.85}
        & \textbf{0.15}
        & \textbf{0.56} \\
        \bottomrule
    \end{tabular*}

    \vspace{-6pt}
\end{wraptable}

\noindent\textbf{Real-world results.}
Given resource constraints, we select D4R-ViT for real-world evaluation, as it achieves the highest overall simulation success rate among the four initializations under our method. As shown in Table~\ref{tab:real_robot_results}, our method improves success rates across all four tasks compared with Frozen and HRP. With D4R-ViT initialization and 40 demonstrations per task, it achieves an average success rate of $56.25\%$, a relative improvement of $73.08\%$ over HRP. Our policy more reliably completes pick-and-place and pushing tasks under randomized object placements. In the pick-and-place task, it more frequently locates and picks carrots at unseen positions. In the pushing task, randomly positioned pots are more consistently moved into the target region. However, lid opening remains challenging for all methods. Our setup uses a single camera positioned diagonally behind the robot, without a complementary wrist-camera view. The limited spatial information and partial occlusion of the small lid make localization difficult, and grasps frequently fail.

\par
\endgroup

\subsection{Human-to-Robot Task-Progress Retrieval}

\begin{table}[t]
    \centering
    \small
    \caption{Human-to-robot task-progress retrieval, reported as correct retrievals out of 100 queries per evaluation setting. Higher is better.}
    \label{tab:stage_retrieval}
    \setlength{\tabcolsep}{5pt}
    \renewcommand{\arraystretch}{1.08}
    \begin{tabular*}{\linewidth}{@{\extracolsep{\fill}}lcccc@{}}
        \toprule
        & \multicolumn{2}{c}{In-distribution (RH20T)}
        & \multicolumn{2}{c}{Zero-shot (self-collected)} \\
        \cmidrule(lr){2-3}\cmidrule(lr){4-5}
        Setting & D4R-ViT & R3M-RN50 & D4R-ViT & R3M-RN50 \\
        \midrule
        Frozen & 15/100 & 14/100 & 42/100 & 36/100 \\
        + HRP / HRAlign & 17/100 & 22/100 & 47/100 & 43/100 \\
        \textbf{+ Ours}
        & \textbf{39/100} & \textbf{43/100}
        & \textbf{56/100} & \textbf{61/100} \\
        \bottomrule
    \end{tabular*}
\end{table}

\begin{figure}[!tbp]
    \centering
    \includegraphics[width=\linewidth]{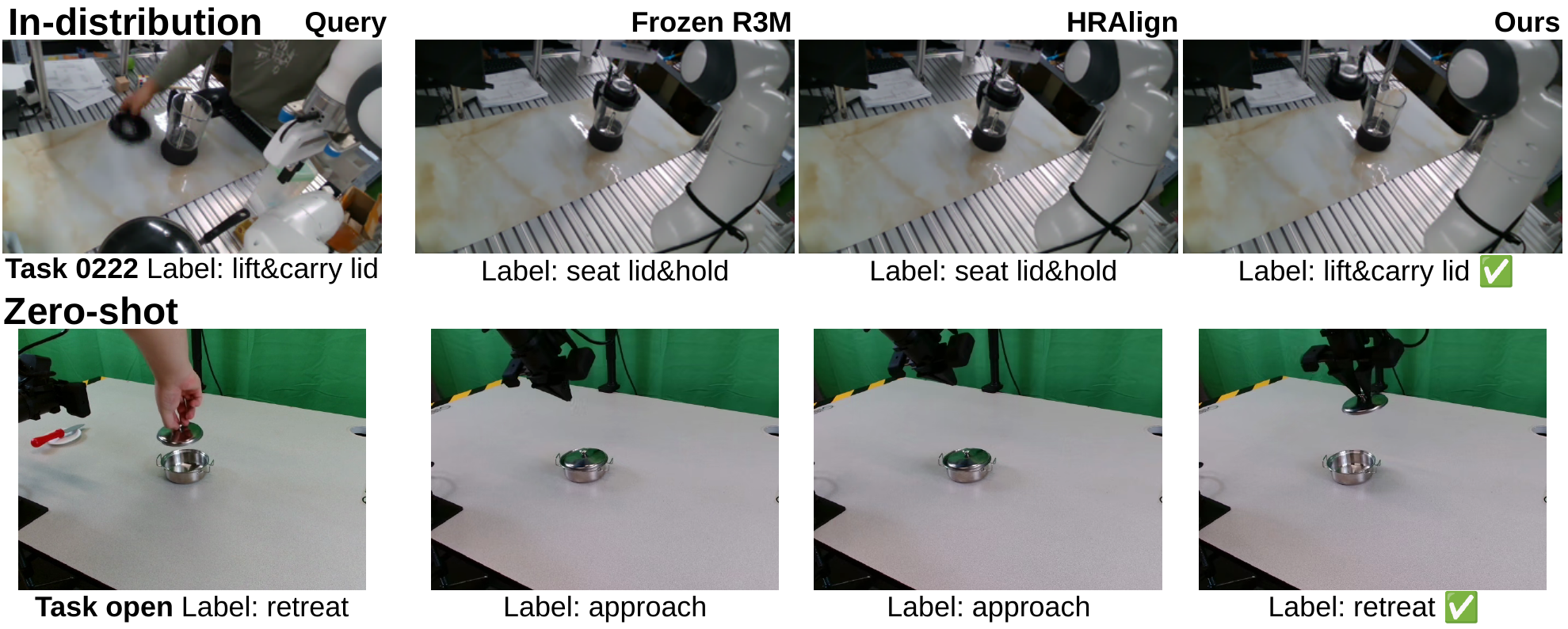}
    \caption{Human-to-robot task-progress retrieval examples from RH20T (top) and the self-collected zero-shot data (bottom). Each row shows a human query and the nearest robot frames retrieved by Frozen R3M, HRAlign, and our method. Green checks indicate correct retrievals.}
    \label{fig:progress_retrieval}
\end{figure}

We evaluate whether human query frames retrieve robot frames at corresponding task progress under in-distribution and zero-shot settings. For in-distribution evaluation, we randomly select ten held-out human--robot video pairs from RH20T without training and randomly sample 100 human query frames across these pairs. For zero-shot evaluation, we use our self-collected data covering four manipulation tasks. For each self-collected task, queries from the five human videos retrieve their nearest-neighbor frames across all five robot videos. We randomly sample another 100 human query frames across this collection. We manually annotate task progress in both human and robot videos, using consistent stage definitions within each task. Each RH20T query retrieves its nearest-neighbor frame from the paired robot video based on visual similarity. A retrieval is considered correct when the query and retrieved frame share the same annotated progress. We report the number of correct retrievals out of 100 queries for each setting. As shown in Table~\ref{tab:stage_retrieval}, our method achieves the highest number of correct retrievals with both visual encoders in both settings. On RH20T, our method yields 22 more correct retrievals than HRP with D4R-ViT and 21 more than HRAlign with R3M-RN50. On the self-collected data, the corresponding gains are 9 and 18 correct retrievals out of 100 queries, respectively. These results provide evidence of task-progress correspondence across human and robot videos. Figure~\ref{fig:progress_retrieval} illustrates retrieval examples from both settings, comparing our method with Frozen R3M and HRAlign. Annotation details and additional examples are provided in Appendix~\ref{appendix:stage_retrieval}.

\subsection{Ablation Studies}
\noindent\textbf{Component Ablation. }
We evaluate four ablations by removing $D$, $Q$, both priors, or the auxiliary frame discrimination loss $\mathcal L_{\mathrm{InfoNCE}}$. As shown in Table~\ref{tab:dq_ablation}, the full model achieves the highest overall success rate with both encoders. Removing $Q$ causes a larger performance drop than removing $D$. Removing both priors produces the largest drop among the component ablations. Removing $\mathcal L_{\mathrm{InfoNCE}}$ reduces success from $18.67\%$ to $16.64\%$ with D4R-ViT and from $17.97\%$ to $16.93\%$ with R3M-RN50. These ablations show that both temporal priors and frame discrimination contribute to downstream robot manipulation.

\noindent\textbf{Alternative Alignment Objectives. }
We compare alignment objectives by replacing $\mathcal L_{\mathrm{TA}}$ with TCC~\citep{Dwibedi_2019_CVPR} or Soft-DTW~\citep{cuturi2017soft}, keeping all other settings unchanged. TCC forms correspondences through soft nearest-neighbor matching followed by a cycle-back step. In our experiments, we observe that visually noisy frames produce diffuse soft-NN weights, which can lead to incorrect cycle-back matches. Soft-DTW requires alignment paths to start and end at the sequence boundaries. When the boundary frames represent different task progress, this constraint can introduce incorrect correspondences. Our alignment objective achieves higher overall success rates than TCC and Soft-DTW with both encoders, as shown in Table~\ref{tab:dq_ablation}.

\begin{table}[t]
    \centering
    \small
    \caption{Ablations and alternatives on the simulation. In the TCC and Soft-DTW variants, $\mathcal{L}_{\mathrm{TA}}$ is replaced by the corresponding alignment loss.}
    \label{tab:dq_ablation}
    \setlength{\tabcolsep}{3pt}
    \renewcommand{\arraystretch}{1.05}
    \begin{tabular*}{\linewidth}{@{\extracolsep{\fill}}lccccccccc@{}}
        \toprule
        \multirow{2}{*}{Variant} & \multirow{2}{*}{$D$} & \multirow{2}{*}{$Q$}
        & \multirow{2}{*}{$\mathcal{L}_{\mathrm{InfoNCE}}$}
        & \multicolumn{3}{c}{D4R-ViT} & \multicolumn{3}{c}{R3M-RN50} \\
        \cmidrule(lr){5-7} \cmidrule(lr){8-10}
        & & & & L1 $\uparrow$ & L2 $\uparrow$ & Avg. $\uparrow$
              & L1 $\uparrow$ & L2 $\uparrow$ & Avg. $\uparrow$ \\
        \midrule
        Full & \checkmark & \checkmark & \checkmark
        & \textbf{18.05 {\tiny (0.52)}} & \textbf{19.47} {\tiny (0.50)} & \textbf{18.67 {\tiny (0.49)}}
        & \textbf{16.82 {\tiny (1.24)}} & \textbf{19.47 {\tiny (0.19)}} & \textbf{17.97 {\tiny (0.67)}} \\
        w/o $\mathcal{L}_{\mathrm{InfoNCE}}$ & \checkmark & \checkmark & --
        & 16.15 {\tiny (0.66)} & 17.27 {\tiny (1.15)} & 16.64 {\tiny (0.54)}
        & 16.31 {\tiny (0.50)} & 17.73 {\tiny (0.19)} & 16.93 {\tiny (0.22)} \\
        w/o $D$ & -- & \checkmark & \checkmark
        & 14.56 {\tiny (0.77)} & 18.53 {\tiny (0.38)} & 16.29 {\tiny (0.43)}
        & 15.69 {\tiny (1.26)} & 17.07 {\tiny (0.82)} & 16.29 {\tiny (0.91)} \\
        w/o $Q$ & \checkmark & -- & \checkmark
        & 13.74 {\tiny (0.77)} & 17.87 {\tiny (0.50)} & 15.54 {\tiny (0.46)}
        & 13.13 {\tiny (0.38)} & 17.47 {\tiny (0.68)} & 15.01 {\tiny (0.36)} \\
        w/o $D$ \& $Q$ & -- & -- & \checkmark
        & 11.18 {\tiny (0.29)} & 15.20 {\tiny (0.65)} & 12.93 {\tiny (0.33)}
        & 11.79 {\tiny (0.52)} & 15.47 {\tiny (0.38)} & 13.39 {\tiny (0.38)} \\
        \midrule
        Soft-DTW & -- & -- & \checkmark
        & 13.44 {\tiny (0.52)} & 17.60 {\tiny (0.57)} & 15.25 {\tiny (0.22)}
        & 13.95 {\tiny (1.02)} & 16.53 {\tiny (0.33)} & 15.07 {\tiny (0.59)} \\
        TCC & -- & -- & \checkmark
        & 13.13 {\tiny (0.66)} & 14.27 {\tiny (1.15)} & 13.62 {\tiny (0.54)}
        & 12.41 {\tiny (0.77)} & 13.07 {\tiny (1.61)} & 12.70 {\tiny (1.11)} \\
        \bottomrule
    \end{tabular*}
\end{table}

\subsection{Additional Analysis}

\begin{figure}[!tbp]
    \centering
    \begin{minipage}[t]{0.46\textwidth}
        \vspace{0pt}
        \centering
        \begin{minipage}[c][2.3cm][c]{\linewidth}
            \centering
            \includegraphics[width=\linewidth]{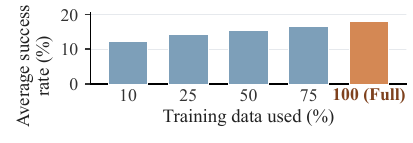}
        \end{minipage}
        {\small (a) Data efficiency}
    \end{minipage}\hfill
    \begin{minipage}[t]{0.51\textwidth}
        \vspace{0pt}
        \centering
        \begin{minipage}[c][2.3cm][c]{\linewidth}
            \centering
            \small
            \setlength{\tabcolsep}{3pt}
            \renewcommand{\arraystretch}{1.08}
            \begin{tabular*}{\linewidth}{@{\extracolsep{\fill}}lcccc@{}}
                \toprule
                Setting & Params. $\downarrow$ & L1 $\uparrow$ & L2 $\uparrow$ & Avg. $\uparrow$ \\
                \midrule
                HRA + Ad. & 1.58M & 11.59 & 12.80 & 12.12 \\
                Ours + Ad. & 1.58M & 14.13 & 15.53 & 14.74 \\
                Ours + BN & \textbf{53.1K} & \textbf{16.82} & \textbf{19.47} & \textbf{17.97} \\
                \bottomrule
            \end{tabular*}
        \end{minipage}
        {\small (b) Adaptation strategies}
    \end{minipage}
    \caption{Data efficiency and adaptation strategies with R3M-RN50. (a) Overall average success rate across L1 and L2 versus the fraction of paired human--robot pre-training data.  (b) Late-adapter (Ad.) versus BN-affine (BN) fine-tuning. Params.: trainable parameters; HRA: HRAlign. All success rates are in percent.}
    \label{fig:efficiency_adaptation}
\end{figure}

\noindent\textbf{Comparison of Encoder Adaptation Strategies.}
To evaluate whether our alignment method also benefits adapter-based fine-tuning, we adopt the late-adapter configuration used by HRAlign. We compare our complete alignment method with reproduced HRAlign using the same R3M initialization, downstream RVT-2 configuration, training data, hyperparameters, and number of training iterations. As shown in Figure~\ref{fig:efficiency_adaptation}(b), our method increases the overall success rate from $12.12\%$ to $14.74\%$ under late-adapter fine-tuning, corresponding to a relative improvement of $21.62\%$. This result indicates that the benefits of our alignment method extend beyond normalization-affine fine-tuning. With our alignment method, updating only 53.1K BN-affine parameters further increases the success rate from $14.74\%$ to $17.97\%$, a relative improvement of $21.91\%$, while the late adapter uses approximately $29.8\times$ as many trainable parameters.

\noindent\textbf{Data Efficiency of Human-Robot Pair Pre-training.}
We evaluate the effect of paired human--robot pre-training data size on downstream performance using R3M-RN50, reporting the overall average success rate across L1 and L2. We vary only the fraction of paired pre-training data while keeping other configurations fixed. As shown in Figure~\ref{fig:efficiency_adaptation}(a), the success rate increases monotonically with the fraction of training data, from $12.28\%$ with $10\%$ of the data to $17.97\%$ with the full dataset. Using $75\%$ of the data achieves a success rate of $16.48\%$. Given the improvement from $75\%$ to $100\%$ of the paired data, we expect further gains in downstream performance from expanding the pre-training dataset.

\section{Related Work}\label{sec:work}

R3M~\citep{pmlr-v205-nair23a} learns reusable visual representations from human videos, while DecisionNCE~\citep{pmlr-v235-li24cr} captures task progression through sampled frame transitions. AcTOL~\citep{zhang2026provable} further introduces temporal ordering and smoothness constraints. Complementary approaches emphasize spatial representations: MVP~\citep{radosavovic2023real} uses masked patch reconstruction, D4R~\citep{pmlr-v229-dasari23a} examines pretraining data, and SlotMIM~\citep{wen2025data} emphasizes object-centric features. These methods provide useful initializations but do not explicitly establish human-robot video correspondences. For human-to-robot transfer, VRB~\citep{bahl2023affordances} predicts contact points and post-contact trajectories to guide manipulation, while HRP~\citep{Srirama-RSS-24} adapts visual representations using hand-object interaction cues. SWIM~\citep{mendonca2023structured}, DexWM~\citep{goswami2025world}, and DreamDojo~\citep{gao2026dreamdojo} instead learn proxy actions from human videos and adapt them to robot action spaces. We focus on cross-domain visual correspondences rather than directly transferring action distributions. Closest to our setting, HRAlign~\citep{Zhou_2025_CVPR} adapts pretrained models using paired human-robot videos, with task instructions guiding shared semantics for video-level alignment. In its limitations, HARP-VLA~\citep{zhu2026harp} highlights execution-rate differences and redundant frames, emphasizing that robust transfer depends on reliable temporal alignment and shared correspondences. TCC~\citep{Dwibedi_2019_CVPR} uses soft nearest-neighbor matching and cycle consistency to encourage frames to return to their original temporal positions. Soft-DTW~\citep{cuturi2017soft} provides a differentiable objective over monotonic warping paths, permitting one-to-many and many-to-one matches, while LAV~\citep{Haresh_2021_CVPR} applies temporal alignment to representation learning. VAVA~\citep{Liu_2022_CVPR} combines optimal transport with temporal priors to accommodate action-order variations and handle unmatched background and redundant frames. Extending temporal alignment across embodiments, XIRL~\citep{pmlr-v164-zakka22a} learns task-progress representations through temporal cycle consistency and uses them to define reinforcement learning rewards.
Inspired by VAVA and XIRL, FLTA uses frame-level temporal correspondences to adapt pretrained visual encoders for downstream manipulation. We refine the video-level semantic alignment used by HRAlign through soft correspondence targets guided by $D$ and local temporal order regularization through $Q$.

\section{Conclusion and Limitations}
We presented a frame-level temporal alignment framework for adapting pretrained visual encoders to robotic manipulation using paired human and robot videos. The global progress prior $D$ and local temporal order prior $Q$ guide correspondence learning, supported by an auxiliary frame discrimination loss. Adaptation updates only the affine parameters of normalization layers and requires no frame-level correspondence annotations. Experiments on AGNOSTOS show higher zero-shot success rates than HRP and HRAlign across four encoder initializations. Real-world experiments with D4R-ViT also show improved manipulation success compared with Frozen and HRP. Task-progress retrieval provides additional evidence of learned cross-domain frame correspondences.

\noindent\textbf{Limitations of Real-World Transfer.}
Precise manipulation remains challenging with few robot demonstrations and a single camera view. With 40 demonstrations per task, the training data may not cover the range of object positions encountered during deployment. Small objects are also difficult to localize from a single external camera and can become occluded during manipulation, contributing to grasping failures. We plan to collect larger and more diverse robot demonstration datasets and incorporate complementary camera views into ACT.

\noindent\textbf{Quality of Paired Videos.}
In some RH20T videos, shifts in camera viewpoint and objects moving out of frame make key parts of the manipulation difficult to observe. When key actions are not visible, visual similarity becomes less reliable for matching task progress across human and robot videos, making temporal alignment more difficult. Future work will explore more informative signals shared by human and robot demonstrations to reduce reliance on visual similarity for temporal alignment.

\section{Impact Statement}
We employed large language models (LLMs) as tools for grammatical refinement, polishing the precision of word choice and the appropriateness of content expression. LLMs enhanced grammatical accuracy, textual clarity, and lexical selection. No aspects of the research design, implementation process, data analysis, or conceptual development relied on LLMs. This work studies the adaptation of visual representations for robotic manipulation using paired human-robot videos. The approach may support more effective reuse of human demonstrations for robot learning.

\begingroup
\small
\setlength{\bibsep}{4pt}
\catcode`，=\active
\def，{,}
\bibliography{references}
\bibliographystyle{references}
\endgroup

\clearpage
\appendix
\section{Appendix}
\subsection{Temporal Differences in Paired Human--Robot Videos}
\label{app:temporal_differences}

\begin{figure}[H]
    \centering
    \includegraphics[width=\textwidth]{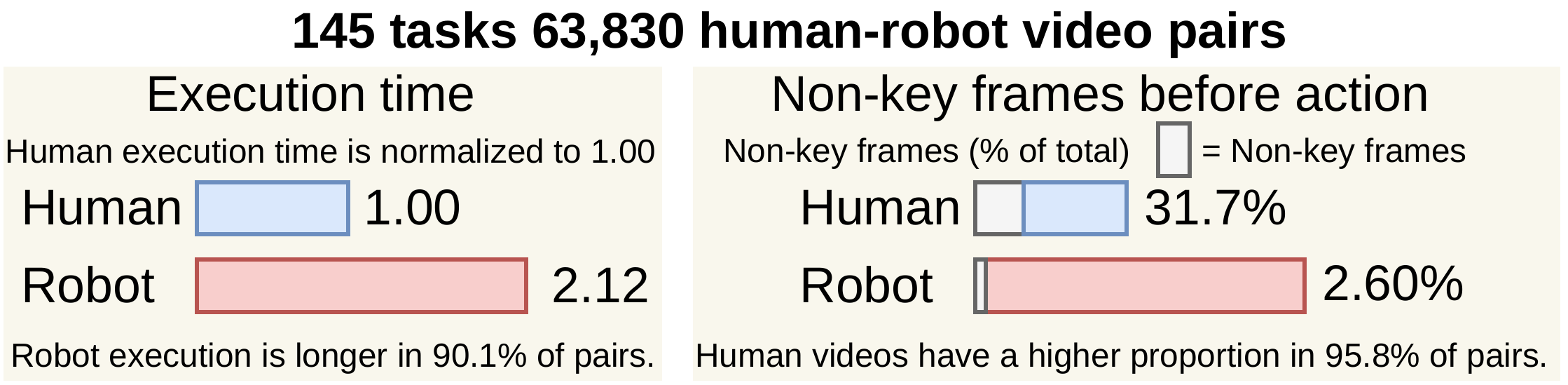}
    \caption{Temporal differences in paired human--robot videos. Left: median Robot-to-Human execution-time ratio, with Human time normalized to 1. Right: mean proportion of non-key frames before action execution.}
    \label{fig:temporal difference}
\end{figure}

Our cleaned RH20T dataset~\citep{fang2024rh20t} contains 63,830 human--robot video pairs across 145 tasks. As shown in Figure~\ref{fig:temporal difference}, Robot execution is longer in 90.1\% of pairs, with a median Robot-to-Human execution-time ratio of 2.12. The mean proportion of non-key frames before action execution is 31.7\% for Human videos and 2.6\% for Robot videos. Human videos have a higher proportion in 95.8\% of pairs. Both differences hold across all 145 tasks when evaluated using within-task medians.

\subsection{Notation}\label{app:notation}

Table~\ref{tab:core_notation} lists the core notation used in the proposed method.

\begin{table}[H]
    \centering
    \small
    \caption{Core notation used in the proposed method.}
    \label{tab:core_notation}
    \renewcommand{\arraystretch}{1.08}

    \begin{tabularx}{\linewidth}{@{}l>{\raggedright\arraybackslash}X@{}}
        \toprule
        Symbol
        & Meaning \\
        \midrule

        $\mathcal H,\mathcal R$
        & Paired human and robot videos. \\

        $f_\theta$
        & Visual encoder with parameters $\theta$; only normalization-affine
          parameters are updated during adaptation. \\

        $\mathcal A_1,\mathcal A_2$
        & Augmentation operators for the two branches. \\

        $q_n^X,k_n^X$
        & Representations of the same frame index $n$ under the two
          augmentations, for domain $X\in\{\mathcal H,\mathcal R\}$. \\

        $\mathcal T,\ |\mathcal T|$
        & Sampled temporal index set and number of sampled frames per video. \\

        $n,N$
        & Frame-instance index and total number of sampled frames per domain. \\

        $t,s$
        & Human frame index $t$ and robot frame index $s$. \\

        \midrule

        $S$
        & Cross-domain visual similarity matrix. \\

        $P$
        & Predicted correspondence matrix. \\

        $D$
        & Global progression cost matrix. \\

        $T$
        & Soft correspondence target constructed from $S$ and $D$. \\

        $Q$
        & Local transition cost matrix. \\

        $\rho,\ \epsilon$
        & Global progression prior weight and Sinkhorn temperature,
          respectively. \\

        \midrule

        $\mathcal R_Q(P)$
        & Regularization term that penalizes backward transitions
          under the predicted correspondences. \\

        $\mathcal L_{\mathrm{InfoNCE}}$
        & Intra-domain frame discrimination loss. \\

        $\mathcal L_{\mathrm{TA}}$
        & Cross-domain temporal alignment loss,
          averaged over both directions. \\

        $\mathcal L_{\mathrm{total}}$
        & Weighted sum of $\mathcal L_{\mathrm{InfoNCE}}$
          and $\mathcal L_{\mathrm{TA}}$,
          with weights $\lambda_1$ and $\lambda_2$. \\

        \bottomrule
    \end{tabularx}
\end{table}

\subsection{Details of the Bidirectional Alignment Objective}
\label{app:alignment_objective}

The following definitions refer to human-to-robot alignment, with directional superscripts omitted. Each row of $T$ and $P$ gives a distribution over robot frames.

\noindent\textbf{Sinkhorn Normalization.}
We perform 20 Sinkhorn iterations, normalizing rows and then columns to sum to one. A final row normalization gives the soft correspondence target $T$.

\noindent\textbf{Soft-target distillation.}
We minimize the mean row-wise cross-entropy to train $P$ to match $T$:
\begin{equation}
    \operatorname{CE}(T,P)
    =
    -\frac{1}{|\mathcal T|}
    \sum_{t=0}^{|\mathcal T|-1}
    \sum_{s=0}^{|\mathcal T|-1}
    T_{ts}\log P_{ts}.
    \label{eq:alignment_distillation_detail}
\end{equation}
We construct $T$ from visual similarity and the global progression prior, and stop gradients through $T$.

\noindent\textbf{Local transition regularization.}
For consecutive sampled human frames, we penalize backward transitions in their robot correspondences:
\begin{equation}
    \mathcal R_Q(P)
    =
    \frac{1}{|\mathcal T|-1}
    \sum_{t=0}^{|\mathcal T|-2}
    P_{t,:}\,Q\,(P_{t+1,:})^\top.
    \label{eq:local_regularization_detail}
\end{equation}
The two rows of $P$ weight the transition costs in $Q$ by their predicted probabilities. $Q$ penalizes backward transitions but not stays or forward transitions.

\noindent\textbf{Bidirectional averaging.}
We exchange the human and robot sequences to compute the reverse objective, then average the two directional losses:
\begin{equation}
    \mathcal L_{\mathrm{TA}} = \frac{1}{2}\left(\mathcal L_{\mathrm{TA}}^{\mathcal H\rightarrow\mathcal R}+\mathcal L_{\mathrm{TA}}^{\mathcal R\rightarrow\mathcal H}\right).
\end{equation}

\subsection{Frame Discrimination Loss}\label{app:frame_discrimination}
\noindent\textbf{Contrastive Objective. }
We introduce InfoNCE \citep{oord2018representation} to learn representations that can distinguish between different frames. We therefore applied data augmentation techniques before the visual encoder $f_\theta$ maps the frame images. For each frame $x_n^X$ from domain $X\in\{\mathcal H,\mathcal R\}$, two augmentation operators $\mathcal A_1$ and $\mathcal A_2$ produce representations $q_n^X=f_\theta(\mathcal A_1(x_n^X))$ and $k_n^X=f_\theta(\mathcal A_2(x_n^X))$, where $n$ indexes these sampled frames across videos.
Let $N$ denote the total number of sampled frame instances per domain, expressed as $N = (\text{number of video pairs}) \times |\mathcal T|$. We average the directional InfoNCE losses over both augmentation directions and both domains:

\begin{equation}
    \mathcal L_{\mathrm{InfoNCE}}
    =
    \frac{1}{4}
    \sum_{X\in\{\mathcal H,\mathcal R\}}
    \left[
        \ell_{\mathrm{NCE}}(q^X,k^X)
        +
        \ell_{\mathrm{NCE}}(k^X,q^X)
    \right].
    \label{eq:bidirectional_infonce}
\end{equation}
Here, $\ell_{\mathrm{NCE}}$ denotes the directional InfoNCE loss~\citep{oord2018representation}, averaged over the $N$ frames. Different augmented branches of the same frame form positive pairs, while representations from other frames serve as negative samples.
\subsection{Pretraining Details}\label{appendix:pre-training details}

The pretraining settings and hyperparameters provided by HRAlign~\citep{Zhou_2025_CVPR} and HRP~\citep{Srirama-RSS-24} serve as a reference for us. However, due to limited computational resources, all of our pretraining was performed on a single NVIDIA RTX 5090, and our dataset utilized RH20T, a task-level dataset of human-robot pairs~\citep{fang2024rh20t}. With approximately 64K human--robot video pairs, pretraining on the ViT encoder took approximately 14.10 hours, while pretraining on the RN50 encoder took 12.97 hours. Details of the pretraining hyperparameters are shown in Table~\ref{tab:pretraining_hyperparameters} and Table~\ref{tab:pretraining_hyperparameters_clip}. We examine the effect of the number of sampled frames in Appendix~\ref{app:frame_sampling}. The loss weights are set to $\lambda_1=0.5$ and $\lambda_2=1$ for both encoders. We set $(\rho,\epsilon)=(0.25,0.025)$ for R3M-RN50, DecisionNCE-CLIP, and AcTOL-CLIP, and $(0.5,0.05)$ for D4R-ViT. And, appearance perturbation includes color jitter, grayscale, blur, and erasing; all inputs are normalized.

\begin{table}[!htbp]
    \centering
    \small
    \caption{Pretraining configurations for ViT and RN50.}
    \label{tab:pretraining_hyperparameters}
    \setlength{\tabcolsep}{10pt}
    \begin{tabular}{@{}lcc@{}}
        \toprule
        \multicolumn{1}{l}{Configuration} & ViT & RN50 \\
        \midrule
        Training iterations & 40,000 & 40,000 \\
        Optimizer & AdamW & AdamW \\
        Learning rate & $7.5\times10^{-5}$ & $1.0\times10^{-4}$ \\
        Frames per domain & 160 & 160 \\
        Sampled frames & 8 & 8 \\
        Weight decay & 0 & 0 \\
        Loss weights $(\lambda_1,\lambda_2)$ & (0.5,1) & (0.5,1) \\
        Global-prior weight $\rho$ & 0.5 & 0.25 \\
        Sinkhorn temperature $\epsilon$ & 0.05 & 0.025 \\
        Augmentation $\mathcal A_1/\mathcal A_2$
            & Resize / Appearance & Resize / Appearance \\
        \bottomrule
    \end{tabular}
\end{table}
\begin{table}[!htbp]
    \centering
    \small
    \caption{Pretraining configurations for DecisionNCE-CLIP and AcTOL-CLIP.}
    \label{tab:pretraining_hyperparameters_clip}
    \setlength{\tabcolsep}{10pt}
    \begin{tabular}{@{}lcc@{}}
        \toprule
        Configuration & DecisionNCE-CLIP & AcTOL-CLIP \\
        \midrule
        Training iterations & 40,000 & 40,000 \\
        Optimizer & AdamW & AdamW \\
        Learning rate & $1.0\times10^{-4}$ & $1.0\times10^{-4}$ \\
        Frames per domain & 160 & 160 \\
        Sampled frames & 8 & 8 \\
        Weight decay & 0 & 0 \\
        Loss weights $(\lambda_1,\lambda_2)$ & $(0.5,1)$ & $(0.5,1)$ \\
        Global-prior weight $\rho$ & 0.25 & 0.25 \\
        Sinkhorn temperature $\epsilon$ & 0.025 & 0.025 \\
        Augmentation $\mathcal A_1/\mathcal A_2$ & Resize / Appearance & Resize / Appearance \\
        \bottomrule
    \end{tabular}
\end{table}

\subsubsection{Number of Sampled Frames}
\label{app:frame_sampling}

\begin{figure}[!tbp]
    \centering
    \includegraphics[width=0.55\linewidth]{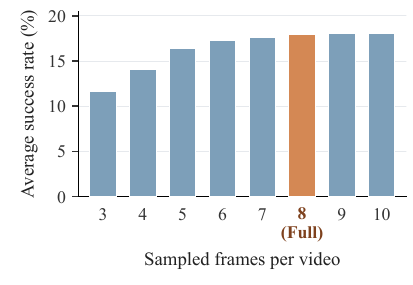}
    \caption{Effect of the number of sampled frames per video on the overall average success rate across L1 and L2 simulation tasks using R3M-RN50. Full denotes the default eight-frame configuration.}
    \label{fig:frame_sampling}
\end{figure}

We evaluate the effect of the number of frames sampled per video during paired human--robot pre-training using R3M-RN50. We vary the number from three to ten and report the overall average success rate across L1 and L2 simulation tasks. As shown in Figure~\ref{fig:frame_sampling}, the success rate increases from $11.65\%$ with three frames to $17.97\%$ with eight frames. Sampling nine and ten frames yields success rates of $18.05\%$ and $18.10\%$, respectively. These correspond to gains of only $0.08$ and $0.13$ percentage points over the eight-frame configuration. In our single-GPU setup, the eight-frame configuration already approaches the GPU memory limit. Sampling nine or ten frames per video requires reducing the batch size from 20 to 16. Under these configurations, adaptation takes $15$--$17$ hours, compared with $12.97$ hours for eight frames. Given the limited performance gains, we adopt eight frames per video as our default configuration, providing a practical trade-off between downstream performance, GPU memory requirements, and adaptation time.

\FloatBarrier
\section{Evaluation Details}
\subsection{Downstream Policy Architectures and Hyperparameters} \label{appendix sec: Downstream Policy Architectures and Hyperparameters}

\noindent\textbf{RLBench 3D Manipulation Policy. }
Table~\ref{tab:downstream_policy_config} summarizes the simulation training configurations. The original RVT-2 architecture consists of two stages. The first stage employs an independent 8-layer attention network for coarse global localization. The second stage employs another independent 8-layer attention network for local, precise prediction. The 8-layer attention network in the second stage is then used to predict the end-effector's position, rotation, gripper state, and collision control state. In contrast, RVT1 has only 1 stage, using a single 8-layer attention network to directly predict the final gripper action from the complete scene, without a local zoom-in or local refinement stage. To evaluate the transfer of pretrained visual representations, we follow HRAlign's lightweight policy design. We extract features from rendered virtual views using the frozen visual encoder and process them with a single policy attention layer. We extended this design to RVT2, but modified only the second stage. We retain the original eight attention layers in the first stage for coarse global localization. The policy was trained on 18 standard RLBench training tasks, using 100 expert demonstrations per task for a total of 1,800 training trajectories. To reduce computational cost and assess the contribution of pre-trained visual encoders under a limited downstream training budget, we train both the ViT-based and RN50-based policies for 25,000 iterations, corresponding to one epoch in our training configuration. We use a global batch size of 3, compared with 192 in the original RVT-2. We evaluate each trained policy checkpoint in three separate runs. Each run uses 25 randomly selected test demonstrations per task, with success determined by the task-defined criteria. We report the mean success rates for Level-1, Level-2, and all 23 tasks, together with the standard deviations across the three evaluation runs.
\begin{table}[!htbp]
    \centering
    \footnotesize
    \caption{Downstream policy training configurations.}
    \label{tab:downstream_policy_config}

    \begin{minipage}[t]{0.52\textwidth}
        \vspace{0pt}
        \centering
        \textbf{Simulation configurations}\\[3pt]
        \setlength{\tabcolsep}{4pt}

        \begin{tabular}{@{}lcc@{}}
            \toprule
            Configuration & ViT & RN50 \\
            \midrule
            Policy architecture & \multicolumn{2}{c}{Modified RVT2} \\
            Stage 1 attention layers & \multicolumn{2}{c}{8} \\
            Stage 2 attention layers & \multicolumn{2}{c}{1} \\
            Attention dimension & \multicolumn{2}{c}{512} \\
            Optimizer & \multicolumn{2}{c}{LAMB} \\
            Base learning rate & \multicolumn{2}{c}{$5\times10^{-5}$} \\
            Weight decay & \multicolumn{2}{c}{$1\times10^{-4}$} \\
            LR schedule & \multicolumn{2}{c}{Cosine} \\
            Warm-up steps & \multicolumn{2}{c}{500} \\
            Batch size/GPU & \multicolumn{2}{c}{3} \\
            Training epochs & \multicolumn{2}{c}{1} \\
            Total training iterations & \multicolumn{2}{c}{25,000} \\
            \bottomrule
        \end{tabular}
    \end{minipage}
    \hfill
    \begin{minipage}[t]{0.43\textwidth}
    \vspace{0pt}
    \centering
    \textbf{Real-world configurations}\\[3pt]
    \footnotesize
    \setlength{\tabcolsep}{4pt}
    \begin{tabular}{@{}p{0.53\linewidth}p{0.37\linewidth}@{}}
        \toprule
        Configuration & ViT \\
        \midrule
        Policy architecture & ACT \\
        Transformer dimension & 512 \\
        Attention heads & 8 \\
        Activation & ReLU \\
        Optimizer & AdamW \\
        Learning rate & $1\times10^{-5}$ \\
        Weight decay & $1\times10^{-4}$ \\
        Training steps & 8K \\
        Batch size & 16 \\
        Proprioception & 7-D joint state \\
        Demonstrations & 40 \\
        Image augmentation & None \\
        \bottomrule
    \end{tabular}
    \end{minipage}
\end{table}

\begin{table}[!htbp]
    \centering
    \scriptsize
    \caption{Per-task success rates (\%) on the 23 unseen AGNOSTOS tasks, averaged over three evaluation runs. Avg.\ reports the average success rate for each level, and Std.\ its standard deviation across runs. DNCE denotes DecisionNCE. $^{*}$ denotes reproduced.}
    \label{tab:agnostos_per_task}
    \setlength{\tabcolsep}{2.5pt}
    \renewcommand{\arraystretch}{1.00}

    \textbf{(a) Level-1 tasks}\\[3pt]

    \resizebox{\textwidth}{!}{%
        \begin{tabular}{@{}llrrrrrrrrrrrrrrr@{}}
            \toprule
            Encoder
            & Method
            & Toilet
            & Knife
            & Fridge
            & Micro.
            & Laptop
            & Phone
            & Seat
            & Lamp Off
            & Lamp On
            & Book
            & Umb.
            & Grill
            & Bin
            & Avg.
            & Std. \\
            \midrule

            \multirow{3}{*}{D4R-ViT}
            & Frozen
            & 0 & 0 & 16.0 & 18.7 & 14.7 & 0
            & 56.0 & 0 & 0 & 0 & 0 & 2.7 & 0
            & 8.31 & 0.50 \\

            & + HRP
            & 0 & 1.3 & 22.7 & 10.7 & 32.0 & 0
            & 45.3 & 2.67 & 2.67 & 0 & 1.3 & 18.7 & 0
            & 10.56 & 1.67 \\

            & \textbf{+ Ours}
            & 0 & 0 & 64.00 & 54.67 & 28.00 & 0
            & 73.33 & 0 & 0 & 0 & 1.33 & 12.00 & 1.33
            & \textbf{18.05} & 0.52 \\

            \midrule

            \multirow{3}{*}{R3M-RN50}
            & Frozen
            & 0 & 1.3 & 29.3 & 32.0 & 13.3 & 0
            & 73.3 & 1.3 & 1.3 & 1.3 & 0 & 10.7 & 0
            & 12.62 & 0.25 \\

            & + HRAlign
            & 0 & 0 & 37.3 & 14.7 & 24.0 & 0
            & 65.3 & 6.7 & 1.3 & 0 & 0 & 1.3 & 0
            & 11.59 & 1.62 \\

            & \textbf{+ Ours}
            & 0 & 6.67 & 64.00 & 33.33 & 46.67 & 0
            & 61.33 & 1.33 & 0 & 0 & 0 & 5.33 & 0
            & \textbf{16.82} & 1.24 \\

            \midrule

            \multirow{3}{*}{DNCE-CLIP}
            & Frozen & 0.00 & 1.33 & 26.67 & 12.00 & 24.00 & 0.00 & 54.67 & 1.33 & 0.00 & 0.00 & 1.33 & 6.67 & 2.67 & 10.05 & 1.24 \\
            & + HRAlign$^{*}$ & 0.00 & 1.33 & 44.00 & 6.67 & 33.33 & 1.33 & 60.00 & 0.00 & 2.67 & 0.00 & 1.33 & 4.00 & 9.33 & 12.62 & 2.06 \\
            & \textbf{+ Ours} & 0.00 & 1.33 & 53.33 & 22.67 & 36.00 & 5.33 & 78.67 & 0.00 & 0.00 & 0.00 & 0.00 & 0.00 & 2.67 & \textbf{15.38} & 0.44 \\

            \midrule

            \multirow{3}{*}{AcTOL-CLIP}
            & Frozen & 0.00 & 1.33 & 28.00 & 21.33 & 29.33 & 0.00 & 60.00 & 0.00 & 0.00 & 0.00 & 0.00 & 2.67 & 4.00 & 11.28 & 1.53 \\
            & + HRAlign$^{*}$ & 0.00 & 2.67 & 36.00 & 28.00 & 13.33 & 0.00 & 64.00 & 0.00 & 0.00 & 0.00 & 0.00 & 1.33 & 9.33 & 11.90 & 1.47 \\
            & \textbf{+ Ours} & 0.00 & 1.33 & 49.33 & 22.67 & 41.33 & 4.00 & 50.67 & 0.00 & 0.00 & 1.33 & 0.00 & 1.33 & 0.00 & \textbf{13.23} & 0.75 \\
            \bottomrule
        \end{tabular}%
    }

    \vspace{5pt}

    \textbf{(b) Level-2 tasks}\\[3pt]

    \resizebox{\textwidth}{!}{%
        \begin{tabular}{@{}llrrrrrrrrrrrr@{}}
            \toprule
            Encoder
            & Method
            & USB
            & Lid
            & Plate
            & Ball
            & Scoop
            & Rope
            & Oven
            & Buzz
            & Plants
            & Charger
            & Avg.
            & Std. \\
            \midrule

            \multirow{3}{*}{D4R-ViT}
            & Frozen
            & 94.7 & 8.0 & 0 & 0 & 0 & 0
            & 10.7 & 0 & 2.7 & 0
            & 11.60 & 0.33 \\

            & + HRP
            & 100.0 & 0 & 0 & 0 & 0 & 0
            & 10.7 & 4.0 & 5.3 & 1.3
            & 12.13 & 0.38 \\

            & \textbf{+ Ours}
            & 93.33 & 85.33 & 0 & 0 & 0 & 0
            & 5.33 & 10.67 & 0 & 0
            & \textbf{19.47} & 0.50 \\

            \midrule

            \multirow{3}{*}{R3M-RN50}
            & Frozen
            & 100.0 & 0 & 0 & 0 & 0 & 0
            & 6.7 & 9.3 & 0 & 1.3
            & 11.73 & 0.50 \\

            & + HRAlign
            & 93.3 & 0 & 0 & 0 & 0 & 0
            & 25.3 & 5.3 & 4.0 & 0
            & 12.80 & 0.33 \\

            & \textbf{+ Ours}
            & 94.67 & 84.00 & 0 & 0 & 0 & 0
            & 6.67 & 6.67 & 2.67 & 0
            & \textbf{19.47} & 0.19 \\

            \midrule

            \multirow{3}{*}{DNCE-CLIP}
            & Frozen & 78.67 & 2.67 & 0.00 & 0.00 & 0.00 & 0.00 & 6.67 & 6.67 & 2.67 & 0.00 & 9.73 & 1.32 \\
            & + HRAlign$^{*}$ & 73.33 & 5.33 & 0.00 & 0.00 & 0.00 & 0.00 & 10.67 & 2.67 & 0.00 & 2.67 & 9.47 & 1.05 \\
            & \textbf{+ Ours} & 68.00 & 69.33 & 4.00 & 2.67 & 0.00 & 0.00 & 9.33 & 2.67 & 1.33 & 4.00 & \textbf{16.13} & 1.00 \\

            \midrule

            \multirow{3}{*}{AcTOL-CLIP}
            & Frozen & 92.00 & 0.00 & 0.00 & 0.00 & 0.00 & 0.00 & 5.33 & 8.00 & 1.33 & 8.00 & 11.47 & 0.19 \\
            & + HRAlign$^{*}$ & 94.67 & 2.67 & 1.33 & 0.00 & 0.00 & 0.00 & 14.67 & 6.67 & 1.33 & 1.33 & 12.27 & 0.19 \\
            & \textbf{+ Ours} & 98.67 & 94.67 & 2.67 & 0.00 & 0.00 & 0.00 & 14.67 & 5.33 & 1.33 & 0.00 & \textbf{21.73} & 0.50 \\
            \bottomrule
        \end{tabular}%
    }
\end{table}

\noindent\textbf{Real-World Behavior Cloning Policy. }
We adopt ACT as the downstream policy architecture, following the official Trossen--LeRobot training pipeline for the WidowX AI. We replace its default ResNet-18 visual encoder with the encoder under evaluation while keeping all other policy and training configurations unchanged. The substituted encoder remains frozen during policy training, allowing us to isolate the effect of its learned visual representations rather than that of end-to-end adaptation. Detailed configurations are provided in Table~\ref{tab:downstream_policy_config}.

\FloatBarrier
\subsection{Human-to-Robot Task-Progress Retrieval}
\label{appendix:stage_retrieval}

We manually annotate task stages in the human and robot videos used for the in-distribution and zero-shot retrieval evaluations. Stage definitions are task-specific and are applied consistently to the human and robot videos of the same task. Frames at shared stage boundaries are assigned to the later stage. The complete frame ranges, stage names and other details are provided in Tables~\ref{tab:stage_frame_ranges} and~\ref{tab:stage_frame_ranges_zero_shot}. For each evaluation setting, we randomly sample 100 human query frames. Each RH20T query retrieves its nearest-neighbor frame from the paired robot video based on visual similarity. For each self-collected task, queries from all five human videos retrieve frames from a shared pool of the five robot videos. A retrieval is counted as correct only when both the query and retrieved frame belong to the same annotated stage. We report the number of correct retrievals out of 100 queries in Table~\ref{tab:stage_retrieval}.

\begin{figure}[!htbp]
    \centering
    \includegraphics[width=\linewidth]{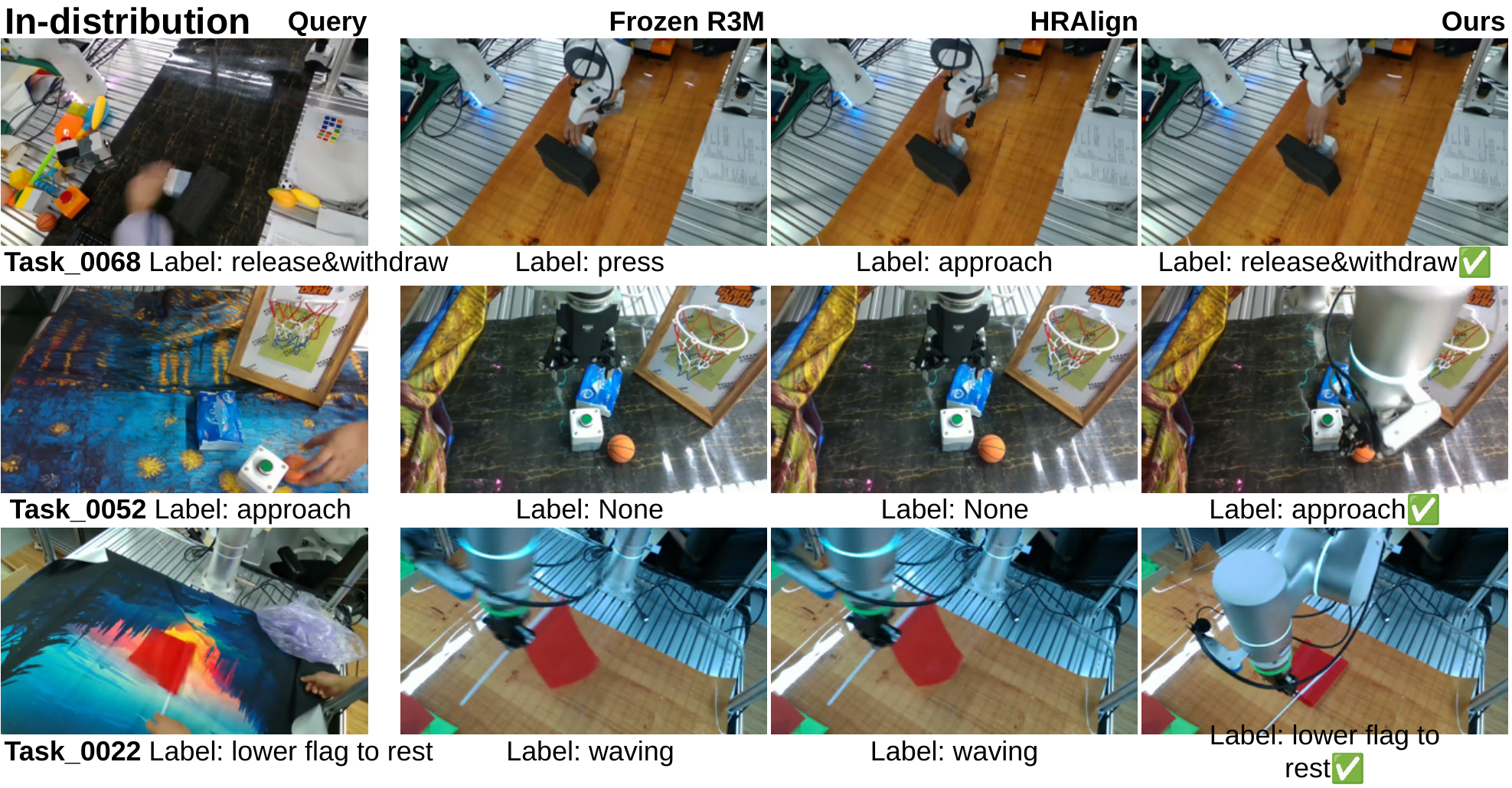}
    \caption{Human-to-robot task-progress retrieval examples from RH20T. Each row shows a human query and the nearest robot frames retrieved by Frozen R3M, HRAlign, and our method. Green checks indicate correct retrievals.}
    \label{fig:retrieval in additional}
\end{figure}
\begin{figure}[!htbp]
    \centering
    \includegraphics[width=\linewidth]{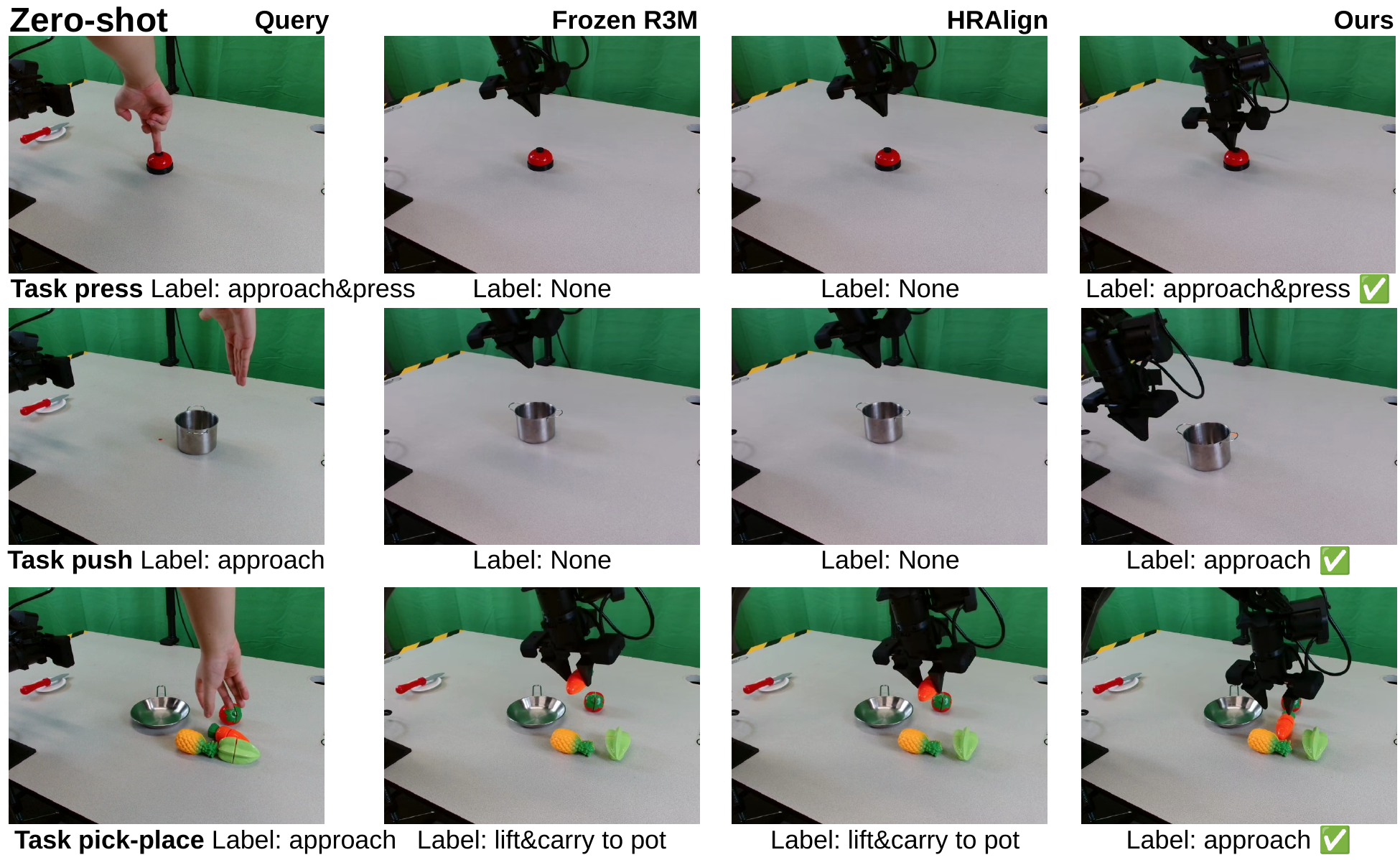}
    \caption{Human-to-robot task-progress retrieval examples from the self-collected zero-shot data. Each row shows a human query and the nearest robot frames retrieved by Frozen R3M, HRAlign, and our method. Green checks indicate correct retrievals. ``None'' denotes frames outside the annotated progress segments.}
    \label{fig:retrieval zero additional}
\end{figure}
Figures~\ref{fig:retrieval in additional} and~\ref{fig:retrieval zero additional} provide additional task-progress retrieval examples from RH20T and the self-collected zero-shot data, respectively. In these selected examples, our method retrieves robot frames from progress segments corresponding to those of the human queries. For the \textit{press} task, we annotate two progress segments: \textit{approach and press} and \textit{release and retreat}. In the example shown in Figure~\ref{fig:retrieval zero additional}, the human query belongs to the \textit{approach and press} segment. Frozen R3M and HRAlign retrieve robot frames outside the annotated action interval, whereas our method retrieves a frame from the corresponding \textit{approach and press} segment. Together with the quantitative results in Table~\ref{tab:stage_retrieval}, these examples provide evidence of task-progress correspondence across human and robot videos and complement the downstream performance results.

\begin{table}[!htbp]
    \centering
    \footnotesize
    \caption{Frame annotations for the ten selected RH20T human--robot pairs. Each task has one human-robot pair. Stage names are task-specific.}
    \label{tab:stage_frame_ranges}
    \setlength{\tabcolsep}{3pt}
    \renewcommand{\arraystretch}{1.08}
    \begin{tabular*}{\textwidth}{@{\extracolsep{\fill}}llcccccc@{}}
        \toprule
        Task & Domain & Outside & Annotated & S1 & S2 & S3 & S4 \\
        \midrule
        \multicolumn{8}{@{}p{\textwidth}@{}}{\textit{task\_0010: S1: approach; S2: grasp and lift; S3: move handset away; S4: retreat}} \\
        task\_0010 & Human & 0--41; 69--85 & 42--68 & 42--46 & 46--53 & 53--64 & 64--68 \\
        task\_0010 & Robot & 0--519 & 520--938 & 520--600 & 600--680 & 680--880 & 880--938 \\
        \addlinespace[3pt]
        \multicolumn{8}{@{}p{\textwidth}@{}}{\textit{task\_0015: S1: approach; S2: lift and aim; S3: striking; S4: return stick and release}} \\
        task\_0015 & Human & 0--43 & 44--69 & 44--47 & 47--55 & 55--64 & 64--69 \\
        task\_0015 & Robot & 0--199 & 200--315 & 200--220 & 220--247 & 247--280 & 280--315 \\
        \addlinespace[3pt]
        \multicolumn{8}{@{}p{\textwidth}@{}}{\textit{task\_0016: S1: approach; S2: lift and aim cue; S3: stroke; S4: ball rolls to rest}} \\
        task\_0016 & Human & 0--39 & 40--90 & 40--46 & 46--68 & 68--82 & 82--90 \\
        task\_0016 & Robot & 0--119 & 120--379 & 120--132 & 132--244 & 244--334 & 334--379 \\
        \addlinespace[3pt]
        \multicolumn{8}{@{}p{\textwidth}@{}}{\textit{task\_0022: S1: approach; S2: grasp and lift; S3: waving; S4: lower flag to rest}} \\
        task\_0022 & Human & 0--23 & 24--70 & 24--32 & 32--41 & 41--66 & 66--70 \\
        task\_0022 & Robot & 0--129 & 130--347 & 130--147 & 147--197 & 197--334 & 334--347 \\
        \addlinespace[3pt]
        \multicolumn{8}{@{}p{\textwidth}@{}}{\textit{task\_0052: S1: approach; S2: grasp and lift; S3: carry and shoot}} \\
        task\_0052 & Human & 0--26; 53--75 & 27--52 & 27--32 & 32--37 & 37--52 & -- \\
        task\_0052 & Robot & 0--49 & 50--280 & 50--86 & 86--188 & 188--280 & -- \\
        \addlinespace[3pt]
        \multicolumn{8}{@{}p{\textwidth}@{}}{\textit{task\_0065: S1: approach and pick; S2: transport brick; S3: place brick}} \\
        task\_0065 & Human & 0--46; 107--134 & 47--106 & 47--61 & 61--90 & 90--106 & -- \\
        task\_0065 & Robot & 0--139 & 140--338 & 140--170 & 170--255 & 255--338 & -- \\
        \addlinespace[3pt]
        \multicolumn{8}{@{}p{\textwidth}@{}}{\textit{task\_0068: S1: approach; S2: press; S3: release and withdraw}} \\
        task\_0068 & Human & 0--95; 124--128 & 96--123 & 96--101 & 101--116 & 116--123 & -- \\
        task\_0068 & Robot & 0--169 & 170--258 & 170--185 & 185--205 & 205--258 & -- \\
        \addlinespace[3pt]
        \multicolumn{8}{@{}p{\textwidth}@{}}{\textit{task\_0092: S1: approach; S2: turn handle; S3: release and withdraw}} \\
        task\_0092 & Human & 0--37; 69 & 38--68 & 38--49 & 49--59 & 59--68 & -- \\
        task\_0092 & Robot & 0--59 & 60--181 & 60--70 & 70--140 & 140--181 & -- \\
        \addlinespace[3pt]
        \multicolumn{8}{@{}p{\textwidth}@{}}{\textit{task\_0105: S1: approach; S2: lift and carry to shelf; S3: release and retreat}} \\
        task\_0105 & Human & 0--67; 192--211 & 68--191 & 68--102 & 102--172 & 172--191 & -- \\
        task\_0105 & Robot & 0--169 & 170--476 & 170--240 & 240--404 & 404--476 & -- \\
        \addlinespace[3pt]
        \multicolumn{8}{@{}p{\textwidth}@{}}{\textit{task\_0222: S1: approach; S2: lift and carry lid; S3: seat lid and hold}} \\
        task\_0222 & Human & 0--22 & 23--137 & 23--48 & 48--120 & 120--137 & -- \\
        task\_0222 & Robot & 0--249 & 250--861 & 250--300 & 300--746 & 746--861 & -- \\
        \bottomrule
    \end{tabular*}
\end{table}

\begin{table}[!htbp]
    \centering
    \scriptsize
    \caption{Frame annotations for the four self-collected zero-shot tasks, with five human and five robot videos per task. Stage names are task-specific.}
    \label{tab:stage_frame_ranges_zero_shot}
    \setlength{\tabcolsep}{3pt}
    \renewcommand{\arraystretch}{1.08}
    \begin{tabular*}{\textwidth}{@{\extracolsep{\fill}}lllccccc@{}}
        \toprule
        Task & Demo & Domain & Outside & Annotated & S1 & S2 & S3 \\
        \midrule
        \multicolumn{8}{@{}p{\textwidth}@{}}{\textit{carrot: S1: approach; S2: lift and carry to pot; S3: release and retreat}} \\
        carrot & 001 & Human & 0--78; 155--164; 169--203 & 79--154; 165--168 & 79--119 & 119--154 & 165--168 \\
        carrot & 001 & Robot & 247--299 & 0--246 & 0--155 & 155--226 & 226--246 \\
        carrot & 002 & Human & 0--86; 169--203 & 87--168 & 87--113 & 113--156 & 156--168 \\
        carrot & 002 & Robot & 253--299 & 0--252 & 0--142 & 142--232 & 232--252 \\
        carrot & 003 & Human & 0--81; 174--218 & 82--173 & 82--108 & 108--157 & 157--173 \\
        carrot & 003 & Robot & 267--299 & 0--266 & 0--138 & 138--246 & 246--266 \\
        carrot & 004 & Human & 0--60; 155--196 & 61--154 & 61--93 & 93--141 & 141--154 \\
        carrot & 004 & Robot & 254--299 & 0--253 & 0--157 & 157--233 & 233--253 \\
        carrot & 005 & Human & 0--50; 145--180 & 51--144 & 51--78 & 78--127 & 127--144 \\
        carrot & 005 & Robot & 285--299 & 0--284 & 0--166 & 166--264 & 264--284 \\
        \addlinespace[3pt]
        \multicolumn{8}{@{}p{\textwidth}@{}}{\textit{open: S1: approach; S2: lift lid and set aside; S3: retreat}} \\
        open & 001 & Human & 0--60; 151--224 & 61--150 & 61--94 & 94--138 & 138--150 \\
        open & 001 & Robot & 233--299 & 0--232 & 0--135 & 135--212 & 212--232 \\
        open & 002 & Human & 0--55; 145--180 & 56--144 & 56--95 & 95--131 & 131--144 \\
        open & 002 & Robot & 248--299 & 0--247 & 0--140 & 140--227 & 227--247 \\
        open & 003 & Human & 0--55; 143--181 & 56--142 & 56--93 & 93--131 & 131--142 \\
        open & 003 & Robot & 253--299 & 0--252 & 0--148 & 148--232 & 232--252 \\
        open & 004 & Human & 0--59; 131--197 & 60--130 & 60--89 & 89--118 & 118--130 \\
        open & 004 & Robot & 260--299 & 0--259 & 0--153 & 153--239 & 239--259 \\
        open & 005 & Human & 0--39; 120--153 & 40--119 & 40--70 & 70--109 & 109--119 \\
        open & 005 & Robot & 273--299 & 0--272 & 0--177 & 177--252 & 252--272 \\
        \addlinespace[3pt]
        \multicolumn{8}{@{}p{\textwidth}@{}}{\textit{press: S1: approach and press; S2: release and retreat}} \\
        press & 001 & Human & 0--43; 93--141 & 44--92 & 44--72 & 72--92 & -- \\
        press & 001 & Robot & 130--159 & 0--129 & 0--119 & 119--129 & -- \\
        press & 002 & Human & 0--51; 96--136 & 52--95 & 52--87 & 87--95 & -- \\
        press & 002 & Robot & 125--159 & 0--124 & 0--114 & 114--124 & -- \\
        press & 003 & Human & 0--39; 88--127 & 40--87 & 40--80 & 80--87 & -- \\
        press & 003 & Robot & 131--159 & 0--130 & 0--120 & 120--130 & -- \\
        press & 004 & Human & 0--56; 116--153 & 57--115 & 57--104 & 104--115 & -- \\
        press & 004 & Robot & 144--159 & 0--143 & 0--133 & 133--143 & -- \\
        press & 005 & Human & 0--77; 128--162 & 78--127 & 78--118 & 118--127 & -- \\
        press & 005 & Robot & 131--159 & 0--130 & 0--120 & 120--130 & -- \\
        \addlinespace[3pt]
        \multicolumn{8}{@{}p{\textwidth}@{}}{\textit{push: S1: approach; S2: push pot to area; S3: release and retreat}} \\
        push & 001 & Human & 0--54; 114--190 & 55--113 & 55--80 & 80--104 & 104--113 \\
        push & 001 & Robot & 195--239 & 0--194 & 0--132 & 132--174 & 174--194 \\
        push & 002 & Human & 0--42; 102--141 & 43--101 & 43--66 & 66--91 & 91--101 \\
        push & 002 & Robot & 194--239 & 0--193 & 0--119 & 119--173 & 173--193 \\
        push & 003 & Human & 0--34; 90--123 & 35--89 & 35--59 & 59--81 & 81--89 \\
        push & 003 & Robot & 213--239 & 0--212 & 0--138 & 138--192 & 192--212 \\
        push & 004 & Human & 0--33; 87--121 & 34--86 & 34--53 & 53--76 & 76--86 \\
        push & 004 & Robot & 195--239 & 0--194 & 0--143 & 143--174 & 174--194 \\
        push & 005 & Human & 0--34; 80--113 & 35--79 & 35--51 & 51--70 & 70--79 \\
        push & 005 & Robot & 205--239 & 0--204 & 0--124 & 124--184 & 184--204 \\
        \bottomrule
    \end{tabular*}
\end{table}
\FloatBarrier
\subsection{Simulation Environment}
RLBench~\citep{james2020rlbench} is a simulated manipulation environment built around the Franka Panda robot and provides multi-view RGB-D observations. Following the widely adopted 18-task protocol established by PerAct~\citep{shridhar2023perceiver} and subsequently used by RVT~\citep{goyal2023rvt} and RVT-2~\citep{goyal2024rvt}, a single language-conditioned policy is trained and evaluated on the same 18 task categories, comprising 249 task variations. This protocol primarily evaluates generalization to variations within seen task categories, rather than zero-shot generalization to entirely unseen tasks. We additionally follow the AGNOSTOS benchmark introduced with X-ICM~\citep{zhou2026exploring}. AGNOSTOS retains the standard 18 RLBench tasks for downstream policy training but evaluates the trained policy on 23 task categories that are absent from the training set, as summarized in Table~\ref{tab:agnostos_tasks}. Level-1 contains 13 unseen tasks that share partial object or action semantics with the training tasks, thereby evaluating transfer through recombinations of familiar semantics. Level-2 contains 10 tasks with previously unseen object and action semantics, requiring stronger cross-task zero-shot generalization. AGNOSTOS therefore evaluates whether a policy can transfer and recombine knowledge acquired from demonstrations of seen tasks to unseen tasks for which no downstream demonstrations are provided. We follow the downstream training and evaluation protocol of RVT and RVT-2~\citep{goyal2023rvt,goyal2024rvt}.
\begin{table}[!htbp]
    \centering
    \small
    \caption{Seen training and unseen testing task categories in the AGNOSTOS benchmark. The 23 testing categories are \textbf{disjoint} from the 18 training categories.}
    \label{tab:agnostos_tasks}
    \setlength{\tabcolsep}{5pt}
    \renewcommand{\arraystretch}{1.08}
    \begin{tabular}{@{}lcp{0.70\linewidth}@{}}
        \toprule
        Split & \# Tasks & Task categories \\
        \midrule
        \textbf{Training} & 18 & Close jar; insert onto square peg; light bulb in; meat off grill; open drawer; place cups; place shape in shape sorter; place wine at rack location; push buttons; put groceries in cupboard; put item in drawer; put money in safe; reach and drag; slide block to color target; stack blocks; stack cups; sweep to dustpan of size; and turn tap. \\
        \midrule
        \textbf{Testing Level-1} & 13 & Put toilet roll on stand; put knife on chopping board; close fridge; close microwave; close laptop lid; phone on base; toilet seat down; lamp off; lamp on; put books on bookshelf; put umbrella in umbrella stand; open grill; and put rubbish in bin. \\
        \addlinespace
        \textbf{Testing Level-2} & 10 & Take USB out of computer; take lid off saucepan; take plate off colored dish rack; basketball in hoop; scoop with spatula; straighten rope; turn oven on; beat the buzz; water plants; and unplug charger. \\
        \bottomrule
    \end{tabular}
\end{table}

\FloatBarrier
\subsection{Real World Environment}

\begin{figure}[!htbp]
    \centering
    \includegraphics[width=\linewidth]{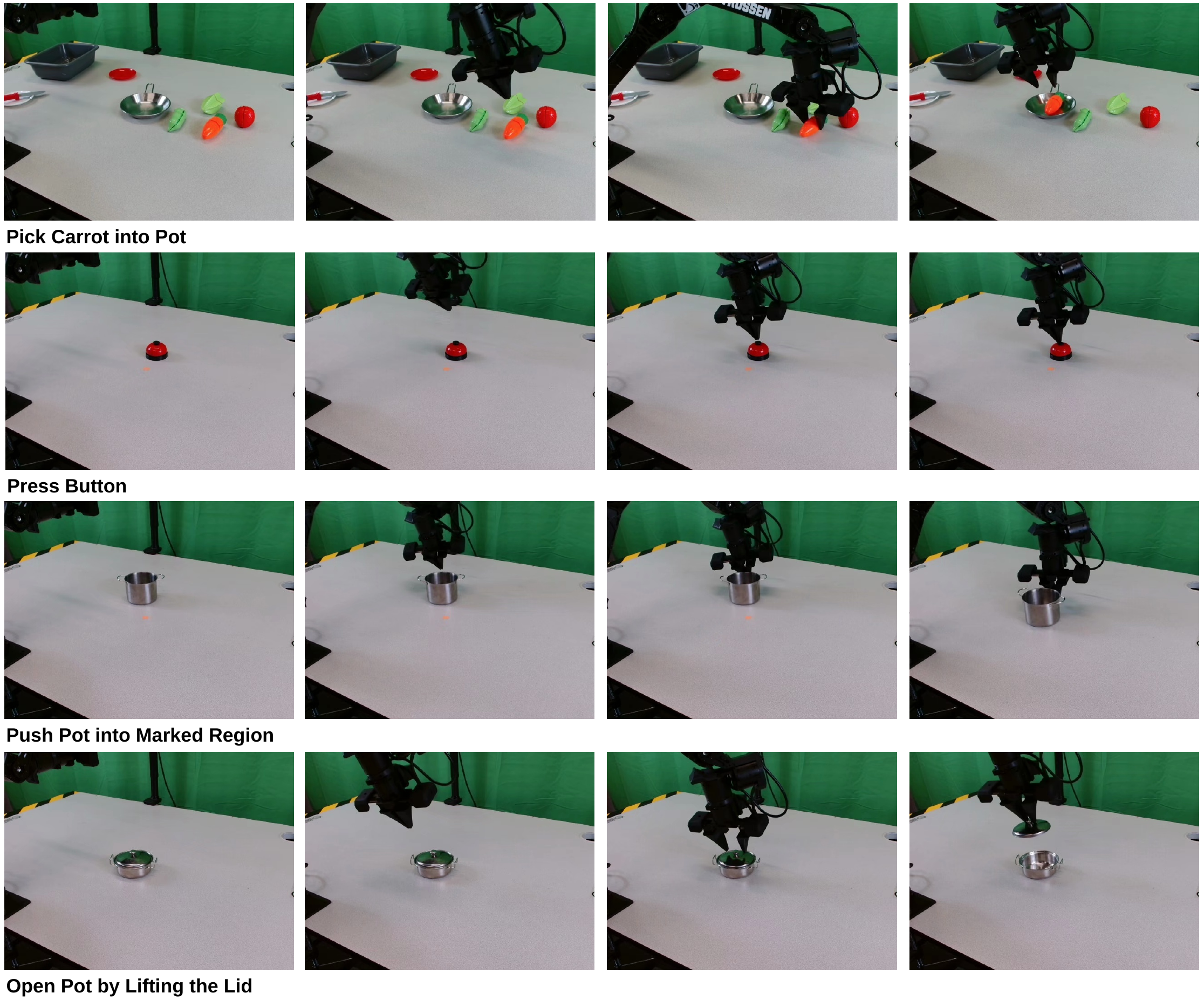}
    \caption{Real-world manipulation tasks evaluated using the Trossen WidowX AI arm.}
    \label{fig:real_world_tasks}
\end{figure}

We conduct real-world experiments using a Trossen WidowX AI arm. The policy state is a seven-dimensional vector comprising six joint angles and a scalar gripper state. At a control frequency of 20 Hz, the policy predicts seven-dimensional absolute targets for the six robot joints and the gripper. These targets are executed using Trossen's original low-level controller and official drivers.

Visual observations are captured using a fixed external Logitech C920 camera positioned above and behind the robot. This configuration provides a third-person, approximately top-down view of the workspace. The camera records RGB images at a resolution of $640\times480$ and a frame rate of 20 Hz. We do not use depth observations, a wrist-mounted camera, or image augmentation.

As illustrated in Figure~\ref{fig:real_world_tasks}, we evaluate four tasks: \textit{Pick Carrot into Pot}, \textit{Press Button}, \textit{Push Pot into Marked Region}, and \textit{Open Pot by Lifting the Lid}. These tasks cover four manipulation skills: pick-and-place, button pressing, goal-directed pushing, and object opening~\citep{yu2020meta,james2020rlbench}. For each task, we collect 40 demonstrations through manual teleoperation, independently randomizing the position of the task-relevant object for each demonstration. During evaluation, each policy is tested over 20 episodes per task, with the object position resampled for every episode.

\end{document}